\documentclass{article}

\usepackage[preprint,nonatbib]{neurips_2026}
\usepackage[utf8]{inputenc} 
\usepackage[T1]{fontenc}    
\usepackage{float}
\usepackage[compatibility=false]{caption}
\usepackage{subfigure}
\usepackage{colortbl}
\usepackage{graphicx}
\usepackage{comment}

\usepackage{graphicx}
\usepackage{colortbl}
\usepackage{comment}
\usepackage{algorithm}
\usepackage{algpseudocode}

\usepackage{hyperref}       
\usepackage{url}            

\usepackage{booktabs}       
\usepackage{amsfonts}       
\usepackage{nicefrac}       
\usepackage{microtype}      
\usepackage{xcolor}         
\usepackage{booktabs}
\usepackage{multirow}
\usepackage{pifont}
\usepackage[table]{xcolor}
\definecolor{lightgray}{gray}{0.9}
\newcommand{\cmark}{\ding{51}} 
\newcommand{\xmark}{\ding{55}} 
\usepackage{float}
\usepackage{amsmath}

\hypersetup{
    colorlinks=true,
    citecolor=blue,   
    linkcolor=blue,   
    urlcolor=blue     
}
\title{SMI: Efficient Self-Supervised Learning via Mutual-Information-Inspired Dependency Optimization}

\author{
Pritam Mishra$^{1}$ \quad
Coloma Ballester$^{1}$ \quad
Dimosthenis Karatzas$^{2}$ \\
\vspace{0.5em}
$^{1}$ Universitat Pompeu Fabra, Barcelona, Spain \\
$^{2}$ Universitat Autònoma de Barcelona, Barcelona, Spain
}

\begin{document}

\maketitle

\begin{abstract}
Self-supervised learning (SSL) has achieved remarkable representation learning performance, but many existing methods rely on large batch sizes, memory banks, momentum encoders, or global synchronization mechanisms that substantially increase computational cost and training complexity. In this work, we propose Semantic Mutual Information (SMI), a lightweight self-supervised objective derived from a mutual-information-inspired dependency formulation under Gaussian assumptions. Unlike conventional correlation matching objectives that operate on high-dimensional feature correlation matrices, SMI performs optimization on a sample-level dependency matrix through a nonlinear transformation of pairwise correlations. This formulation induces distinct optimization dynamics that emphasize strongly dependent semantic pairs while maintaining representation diversity. Experimental results on ImageNet using a ResNet-50 backbone demonstrate that SMI achieves competitive linear evaluation performance relative to state-of-the-art SSL approaches while substantially reducing computational complexity. Across multiple low-resource benchmarks, SMI consistently improves transfer performance over Barlow Twins, particularly on fine-grained datasets. Furthermore, analyses of optimization dynamics and representation geometry suggest improved alignment--redundancy balance, greater feature diversity, and more spatially localized semantic representations. These results indicate that nonlinear dependency optimization provides an effective and computationally efficient alternative to conventional correlation-based self-supervised learning objectives.
\end{abstract}

\section{Introduction}

Self-supervised learning (SSL) has emerged as a powerful paradigm for learning transferable visual representations from large-scale unlabeled data. Recent contrastive approaches achieve remarkable performance by maximizing agreement between different augmented views of the same image while separating representations from other samples in the training batch. Methods such as SimCLR~\cite{chen2020simple} and MoCo~\cite{he2020momentum} demonstrate that instance discrimination can learn representations competitive with supervised pre-training across a wide range of downstream tasks. Beyond image classification, these representations have proven effective in applications including video understanding, video summarization, and sequential representation learning~\cite{trim,trimmer}. Despite their success, contrastive methods typically rely on large batch sizes, memory banks, momentum encoders, or cross-device synchronization to provide sufficient negative sample diversity, substantially increasing computational cost and implementation complexity.

To alleviate these limitations, recent non-contrastive approaches shift the focus from explicit negative-pair discrimination toward redundancy reduction and feature decorrelation~\cite{barlowtwins,vicreg}. Methods such as Barlow Twins~\cite{barlowtwins}, VICReg~\cite{vicreg}, and BYOL~\cite{byol} demonstrate that strong representations can emerge without explicit negatives. Nevertheless, these approaches often depend on large-scale distributed training, global batch normalization, gather layers, or architectural asymmetries to avoid representational collapse~\cite{siamsiam}. Moreover, most redundancy-reduction methods operate directly on high-dimensional feature correlation matrices whose computational cost grows quadratically with representation dimensionality, making efficient large-scale training increasingly challenging.

From an information-theoretic perspective, representation learning can be viewed as maximizing dependency between semantically consistent views of the same sample. Mutual information (MI) provides a principled measure of statistical dependency and captures richer relationships than linear correlation alone. However, exact MI estimation is generally intractable in high-dimensional settings~\cite{poole2019variational,belghazi2018mutual}, while practical approximations such as InfoNCE or MINE often introduce additional optimization complexity, sampling constraints, or auxiliary networks. Consequently, despite its conceptual appeal, developing scalable and computationally efficient MI-inspired objectives for SSL remains an open challenge.

In this work, we introduce \textit{Semantic Mutual Information} (SMI), a lightweight self-supervised objective based on a Gaussian mutual-information-inspired dependency transformation. Rather than directly optimizing correlations, SMI first maps pairwise correlations into a nonlinear dependency space through a closed-form mutual-information-inspired transformation and subsequently performs alignment and redundancy reduction on the resulting dependency matrix. This induces optimization dynamics that differ fundamentally from conventional correlation matching objectives.

A key distinction of SMI is that dependency estimation is performed between samples rather than embedding dimensions. Existing redundancy-reduction methods typically operate on feature-level correlation matrices whose size grows with the embedding dimension. In contrast, SMI constructs a sample-level dependency matrix, resulting in substantially lower computational complexity when high-dimensional projection heads are used. Consequently, SMI combines nonlinear dependency optimization with a computationally efficient sample-space formulation, eliminating the need for memory banks, momentum encoders, gather layers, or global batch normalization.

The term \textit{semantic} in SMI refers to dependency structures that remain consistent across semantically invariant augmentations. In self-supervised learning, aggressive augmentations may occasionally produce positive pairs with limited semantic overlap despite originating from the same image. The nonlinear dependency transformation used by SMI naturally allocates greater optimization emphasis to strongly dependent pairs while reducing sensitivity to weakly dependent pairs. We hypothesize that this behavior encourages learning from semantically consistent views and contributes to improved transfer performance, particularly in fine-grained recognition settings.

We evaluate SMI on ImageNet using a ResNet-50 backbone and demonstrate competitive linear evaluation performance relative to state-of-the-art SSL methods while substantially reducing computational complexity and synchronization requirements. Furthermore, we perform extensive analyses of optimization dynamics and representation geometry, including gradient stability, representation diversity, eigenspectrum statistics, and activation energy maps. Across multiple low-resource benchmarks, SMI consistently improves downstream transfer performance over Barlow Twins, with particularly large gains on fine-grained datasets such as ImageWoof.

Our contributions can be summarized as follows:

\begin{itemize}

\item \textbf{Nonlinear Dependency Optimization:}
We introduce Semantic Mutual Information (SMI), a self-supervised objective that performs alignment and redundancy reduction in a nonlinear dependency space derived from a Gaussian mutual-information-inspired transformation.

\item \textbf{Efficient Sample-Level Formulation:}
Unlike conventional redundancy-reduction methods that operate on feature-level correlation matrices, SMI performs dependency optimization in the sample domain, substantially reducing computational complexity while eliminating the need for momentum encoders, gather layers, memory banks, and global batch normalization.

\item \textbf{Optimization and Representation Analysis:}
We provide theoretical and empirical analyses demonstrating that SMI induces distinct optimization dynamics, improved representation diversity, a more balanced alignment--redundancy tradeoff, and consistently improved transfer performance under low-resource training settings.

\end{itemize}

\section{Proposed Method}\label{sec:proposedmethod}

We propose a self-supervised framework that learns robust and discriminative representations by maximizing mutual information (MI) between noise-invariant paired samples—generated from two distorted views of the same instance—while minimizing MI across other pairs within a batch. At the core of this approach is the Semantic Mutual Information (SMI) loss, tailored for resource-constrained environments and which readily extends to multi-view, multi-modal data. Although our experiments focus on visual data, SMI is generic and can be applied to any  combination of modalities. The following subsections detail the mathematical formulation of the loss and present the key theoretical principles underlying its design.
\begin{figure}[ht]
\begin{center}
\centerline{\includegraphics[width=0.8\textwidth]{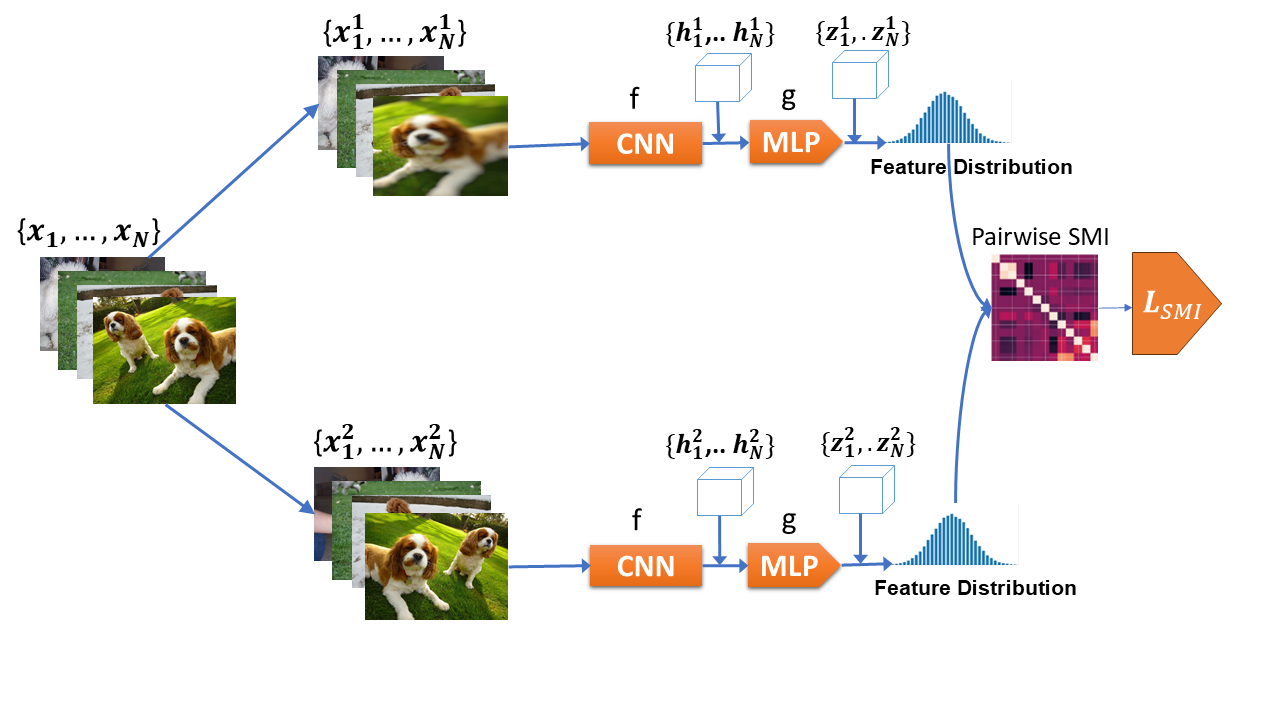}}
\caption{Overview of the proposed method with our SMI loss.}
\label{fig:pipeline}
\end{center}
\end{figure}

\subsection{Gaussian Mutual Information Dependency Transformation}

Mutual information (MI) provides a principled measure of statistical dependency between random variables. However, estimating MI in high-dimensional spaces is generally intractable and often requires variational approximations or negative sampling strategies. To obtain a computationally efficient dependency measure suitable for self-supervised learning, we adopt a Gaussian copula assumption, which yields a closed-form relationship between mutual information and Pearson correlation.

According to Sklar's theorem \cite{sklar1973random}, the dependency structure of a joint distribution can be separated from its marginal distributions through a copula function. Under a Gaussian copula, the dependency between two random variables is fully characterized by their Pearson correlation coefficient $\rho$, defined as

\begin{equation}
\label{eq:rho}
\rho_{X,Y} = \frac{E[XY]-E[X]E[Y]}
{\sqrt{\mathrm{Var}(X)\mathrm{Var}(Y)}}
\end{equation}

For jointly Gaussian variables, mutual information admits the closed-form expression

\begin{equation}
I(X;Y) = -\frac{1}{2}\log(1-\rho^2)
\end{equation}

To ensure numerical stability when $\rho \rightarrow 1$, we introduce a small positive constant $\epsilon$, yielding

\begin{equation}
\label{eq:mi}
I(X;Y)= -\frac{1}{2}\log(1-\rho^2+\epsilon).
\end{equation}

Equation~(\ref{eq:mi}) defines a nonlinear dependency transformation that maps correlation values into a mutual-information-inspired dependency space. Unlike conventional redundancy reduction methods that operate directly on correlations, the proposed SMI objective performs alignment and redundancy reduction on the transformed dependency matrix induced by Eq.~(\ref{eq:mi}). The resulting optimization behavior is analyzed in Section~\ref{ssec:smi_vs_corr}.

\subsection{Proposed Loss leveraging SMI}\label{ssec:SMI}

Let $\{\mathbf{x}_1,\dots,\mathbf{x}_{N}\}\subset\mathcal{I}$ be $N$ samples (a batch, in practice). Let us assume that each sample is described by a vector of size $D$, 
$\mathbf{x}_i\in \mathbb{R}^D$.  For each  $\mathbf{x}_i$, let $t_1$ and $t_2$ two augmentation functions of a set of random transforms $\mathcal{T}$. Now, $\mathbf{x}^1_{i} = t_1(\mathbf{x}_i)$ and  $\mathbf{x}^2_{i} = t_2(\mathbf{x}_i)$ represent two views of $\mathbf{x}_i$. Repeating this process for $i\in\{1,\dots,N\}$, we obtain two sets of $N$ samples  $\{\mathbf{x}^1_1,\dots,\mathbf{x}^1_{N}\}$ and $\{\mathbf{x}^2_1,\dots,\mathbf{x}^2_{N}\}$. Let $\mathbf{h}^v_i=\Phi(\mathbf{x}^v_i)\in \mathbb{R}^M$ be the high-dimensional feature vector obtained by applying an encoder $\Phi$ to each $\mathbf{x}^v_i\in \mathbb{R}^D$, for $v=1,2$. A projection head $g$ maps  $\mathbf{h}^v_i$ to the final embedding vector defined by $\mathbf{z}^v_{i} = g(\mathbf{h}^v_{i})\in \mathbb{R}^K$ of dimension $K<M$, for $v=1,2$ and $i=1,\dots,N$. Figure~\ref{fig:pipeline} displays our  pipeline.

From \eqref{eq:rho} and \eqref{eq:mi} we can compute the MI of $\mathbf{z}^1_i$ and $\mathbf{z}^2_i$ as
$ \mathbf{I}_{1,2} = I(\mathbf{z}^1_i; \mathbf{z}^2_i) = -\frac{1}{2} \log(1 - \rho_{1,2}^2 + \epsilon). $
Similarly, we can compute the MI within the batch of each augmented view, 
$\mathbf{I}_{k,k} = I(\mathbf{z}^k_i; \mathbf{z}^k_i) = -\frac{1}{2} \log(1 - \rho_{k,k}^2 + \epsilon)$, for $k=1,2$. 
Since $\mathbf{I}_{1,2}$ is a symmetric matrix, the diagonal elements can be expressed as
$ D_{1,2} = \left\{ (\mathbf{I}_{1,2})_{m,n} \mid m = n \right\}$. 
The off-diagonal elements of $\mathbf{I}_{1,2}$ , $\mathbf{I}_{1,1}$ and $\mathbf{I}_{2,2}$ can be expressed as
 $ O_{k,l} = \left\{ (\mathbf{I}_{k,l})_{m,n} \mid m \neq n \right\}$, for $k,l=1,2$. 
 
We now define the \emph{on diagonal loss} from $D_{1,2}$ as
\[
L^{D}_{1,2} = \sum \log \left( \cosh \left( D_{1,2} - 1 \right) \right).
\]
The \emph{off diagonal losses} from the off diagonal elements are  defined as
\begin{equation}
L^{O}_{k,l} = \sum \log \left( \cosh \left( O_{k,l} + 0.06 \right) \right) , \quad \text{for} \, k,l=1,2.
\end{equation}
Finally, our SMI loss is  defined as
\begin{equation}\label{eq:finalloss}
    \mathcal{L}_{\text{SMI}}= L^{D}_{1,2} + \lambda(L^{O}_{1,2} + L^{O}_{1,1} + L^{O}_{2,2}).
\end{equation}

\subsection{Difference Between SMI and Correlation Objectives}
\label{ssec:smi_vs_corr}

Although SMI is derived from Pearson correlation under Gaussian assumptions, it differs fundamentally from conventional correlation matching and redundancy reduction objectives. Existing methods such as Barlow Twins and VICReg operate directly on feature-level correlation statistics, enforcing alignment and decorrelation within the original correlation space. In contrast, SMI introduces a nonlinear dependency transformation prior to optimization, resulting in a fundamentally different representation of pairwise relationships and consequently different optimization dynamics.

Given two feature vectors, SMI first computes their Pearson correlation coefficient $\rho$ and maps it into a dependency measure using the Gaussian mutual-information-inspired transformation

\[
M(\rho)
=
-\frac{1}{2}\log(1-\rho^2+\epsilon)
\]

where $\epsilon$ is a small constant introduced for numerical stability. Rather than directly optimizing correlation values, SMI performs alignment and redundancy reduction on the transformed dependency matrix induced by $M(\rho)$. Consequently, optimization is carried out in a nonlinear dependency space rather than the original correlation space.

This distinction can be understood by examining the sensitivity of the dependency transformation with respect to the correlation coefficient:

\[
\frac{\partial M(\rho)}
{\partial \rho}
=
\frac{\rho}
{1-\rho^2+\epsilon}
\]

Unlike linear correlation objectives whose sensitivity remains approximately constant across different correlation values, the proposed transformation introduces correlation-dependent sensitivity. As correlation magnitude increases, the dependency measure becomes increasingly responsive to changes in correlation structure. Consequently, highly correlated feature pairs and weakly correlated feature pairs contribute differently to optimization, producing optimization dynamics that cannot be obtained through direct correlation matching alone.

Importantly, the derivative above characterizes the nonlinear dependency transformation rather than the complete SMI loss. The final objective additionally applies alignment and redundancy reduction constraints to the transformed dependency matrix. Therefore, SMI can be interpreted as a two-stage optimization procedure: (i) pairwise correlations are mapped into a mutual-information-inspired dependency space, and (ii) alignment and redundancy reduction are performed within this transformed space.

\begin{figure}[H]
    \centering
    \includegraphics[width=0.8\columnwidth]{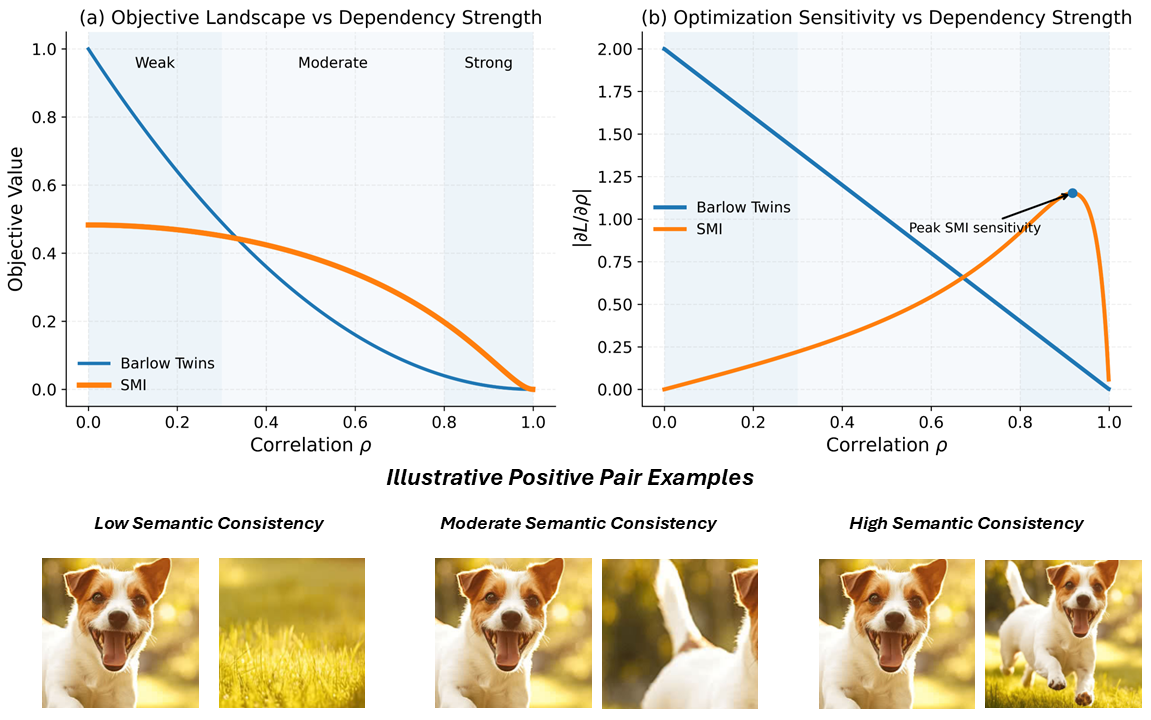}
    \vspace{10pt}
    \caption{Analytical comparison of the optimization behavior induced by Barlow Twins and the proposed SMI objective. \textbf{Left:} Objective value as a function of dependency strength. \textbf{Right:} Optimization sensitivity measured by the gradient magnitude with respect to the correlation coefficient. Unlike direct correlation matching, SMI exhibits correlation-dependent sensitivity, assigning relatively little optimization emphasis to weak dependency regions while focusing optimization effort on moderate-to-high dependency regions. Bottom: illustrative examples of positive pairs exhibiting different levels of semantic consistency.}
    \label{fig:semantic_overlap}
\end{figure}

To better understand the optimization behavior induced
by this formulation, Figure~\ref{fig:semantic_overlap}
compares the objective landscape and optimization
sensitivity of SMI and Barlow Twins as a function of
dependency strength. Figure~\ref{fig:semantic_overlap} illustrates the optimization behavior induced by the proposed objective. Unlike direct correlation matching, SMI exhibits correlation-dependent sensitivity, allocating relatively little optimization emphasis to weak dependency regions while concentrating optimization effort on moderate-to-high dependency regions. This behavior arises from the combination of the nonlinear dependency transformation and the subsequent alignment objective.

One possible implication of this behavior concerns the quality of positive pairs generated by aggressive data augmentations. In self-supervised learning, augmentation strategies such as RandomResizedCrop may occasionally produce views that share limited semantic content despite originating from the same image. Since such pairs are expected to exhibit weaker dependencies, SMI naturally assigns them lower optimization weight compared to positive pairs exhibiting stronger semantic consistency. Consequently, optimization is preferentially focused on views that preserve more consistent semantic structure across augmentations. We hypothesize that this property contributes to the improved transfer performance observed on fine-grained recognition datasets, where preserving semantically informative structures is particularly important.

Beyond the differences in optimization behavior,
SMI also differs structurally from existing
redundancy-reduction methods in the way dependencies
are estimated. A second important distinction lies in the structure of the dependency matrix itself. Methods such as Barlow Twins compute feature-wise correlations between embedding dimensions, producing a correlation matrix of size $K \times K$, where $K$ denotes the embedding dimensionality. Consequently, both memory consumption and computational cost grow quadratically with the representation dimension. In practice, modern redundancy-reduction methods often employ high-dimensional projection heads (e.g., $K=8192$ in Barlow Twins), resulting in increasingly expensive correlation estimation and redundancy reduction as representation dimensionality grows.

In contrast, SMI computes pairwise dependencies between samples, producing a dependency matrix of size $N \times N$, where $N$ denotes the batch size. The proposed objective therefore operates in the sample domain rather than the feature domain. In typical self-supervised learning settings, the embedding dimensionality is substantially larger than the batch size ($K \gg N$), yielding a significantly smaller dependency matrix. For example, with a projection dimension of $K=8192$ and a batch size of $N=256$, Barlow Twins requires an $8192 \times 8192$ correlation matrix, whereas SMI operates on a $256 \times 256$ dependency matrix. This corresponds to more than three orders of magnitude fewer matrix entries, substantially reducing the computational and memory overhead associated with dependency estimation while avoiding large feature-wise redundancy matrices.

From this perspective, the key distinction between SMI and existing redundancy reduction methods is not merely the use of a mutual-information-inspired dependency measure, but the combination of (i) a nonlinear dependency transformation and (ii) sample-level dependency optimization. Rather than directly enforcing redundancy reduction on feature-wise correlations, SMI first maps correlations into a nonlinear dependency space and subsequently performs alignment and redundancy reduction on the resulting sample-level dependency matrix. As demonstrated in Section~\ref{ssec:optimization_dynamics}, this alternative optimization formulation leads to different optimization dynamics, a more balanced alignment--redundancy tradeoff, and improved representation diversity.

\textbf{Implementation Details :}
For ImageNet experiments, we follow the same augmentation pipeline, ResNet-50 architecture, projection head design, and optimization protocol used in Barlow Twins \cite{barlowtwins}. Unlike Barlow Twins and VICReg, SMI does not require Gather Layers or Global Batch Normalization. For the small-scale dataset experiments, identical training hyperparameters are used across all four datasets to ensure a consistent evaluation protocol. Complete implementation details, augmentation parameters, architectures, and optimization hyperparameters are provided in the supplementary material.

\section{Experimental results}\label{sec:experimentalResults}
In this section, we examine both the effectiveness and efficiency of the proposed SMI objective. We first demonstrate competitive representation quality on ImageNet through linear evaluation, then analyze the computational advantages of SMI relative to existing SSL objectives. Next, we study the optimization dynamics and representation diversity induced by SMI, and conclude with extensive evaluations on several low-resource datasets to assess robustness and generalization.
\subsection{ImageNet Linear Evaluation}\label{ssec:lineareval}
We assess the model's performance by training a linear classifier on top of the fixed representations extracted from a ResNet50 model pre-trained using the proposed loss function \eqref{eq:finalloss} on the ImageNet \cite{imagenet} validation set. Our results demonstrate that the proposed method achieves performance comparable to other SOTA approaches, with high efficacy. Table \ref{table:linearevalsota} highlights the performance on the ImageNet \cite{imagenet} validation set, comparing our method to other SOTA approaches.

\begin{table}[tb]
\centering
\small
\caption{\textbf{Top-1 Accuracy on ImageNet\cite{imagenet}:} Linear evaluation on frozen ResNet-50 features, comparing our proposed method with SOTA approaches. \textbf{SN} indicates methods that use shared network weights vs dual ResNet-50 networks (e.g., DINO \cite{dino}, BYOL \cite{byol}). \textbf{MC} refers to multi-crop augmentation, which improved performance in \cite{swav}. The top-3 SOTA methods are ranked as superscripts trained for 300 epochs.}
\vspace{10 pt}
\begin{tabular}{lccccc}
\toprule
Method & Top-1 & Top-5 & SN & MC & Epochs \\
\midrule
\textbf{NTxent$^{Reproduced}$} \cite{chen2020simple}& 51.3 & 76.9 & \cmark & \xmark & 300 \\
\midrule

PIRL \cite{PIRL}& 63.6 & - & \cmark & \xmark & 800\\
SIMCLR \cite{chen2020simple} & 69.3 & 89.0 & \cmark & \xmark & 1000 \\
MoCo v2 \cite{moco_v2} & 71.1 & 90.1 & \xmark & \xmark & 800\\
SIMSIAM \cite{siamsiam}& 71.3 & - & \cmark & \xmark & 800 \\
SWAV \cite{swav}& 71.8 & - & \cmark & \xmark & 800\\
MoCo \cite{he2020momentum} & 60.6 & - & \xmark & \xmark & 200 \\
\textbf{SimCLR\textsuperscript{3}} \cite{chen2020simple}& 67 & 87.5 & \cmark & \xmark & 300 \\
\textbf{Barlow\textsuperscript{2}} \cite{barlowtwins}& 71.4 & 90.2 & \cmark & \xmark & 300 \\
\textbf{BYOL\textsuperscript{1}} \cite{byol}& 72.5 & 91.6 & \xmark & \xmark & 300 \\
\textbf{DINO\textsuperscript{1}} \cite{dino}& 72.5 & 91.6 & \xmark & \xmark & 300 \\
\textbf{Ours\textsuperscript{2}} & 71.41 & 90 & \cmark & \xmark & 300 \\
\bottomrule
\end{tabular}
\label{table:linearevalsota}
\end{table}

\subsection{Computational Efficiency}
\label{ssec:computation_complexity}
To ensure a fair comparison with other SOTA self-supervised methods, we evaluated GFLOPs from two augmentations of shape $(256, 8192)$ (same as (batch size,feature size)) for each loss function. The results demonstrate that our loss function is significantly more computationally efficient compared to all other SOTA losses. The detailed results are presented in Table \ref{table:complexityonImageNet}. We further highlight that our loss function has significantly fewer negative pair terms compared to all other loss functions since it does not require gather layer, while achieving notably higher top-1 accuracy in linear evaluation than NT-Xent~\cite{chen2020simple} (Table \ref{table:linearevalsota}), and performing on par with other SOTA loss functions.

\begin{table}[H]
  \centering
  \footnotesize
  \caption{\textbf{Computational complexity comparison of state-of-the-art SSL objectives and the proposed SMI loss.}GFLOPs denotes the computational complexity of the loss computation during forward propagation for an effective batch size of 2048 distributed across 8 GPUs. \textbf{Off-Diag Entries} denotes the number of pairwise non-target interactions participating in the optimization objective. For contrastive methods, these correspond to negative pairs, whereas for redundancy reduction methods they correspond to off-diagonal correlation or dependency entries. GBN denotes the use of Global Batch Normalization, GL denotes Gather Layer.
}
  \label{table:complexityonImageNet}
  \begin{center}
  \begin{tabular}{lccccc}
    \toprule
    \textbf{Loss} & \textbf{GFLOPs}$\downarrow$ & \textbf{Off-Diag Entries}$\downarrow$ &  \textbf{GBN} & \textbf{GL} & \textbf{Epochs}\\
    \midrule
    Barlow & 137.43 & $8192\times8191$ & \cmark & \cmark & $1000$\\
    VIC-Reg & 274.87 & $2(8192\times8191)$ & \cmark & \cmark & $1000$\\
    Nt-xent & 103.07 & $3(4096\times4095)$ & \cmark & \cmark & 1000\\
    \textbf{SMI} & \textbf{12.88} & $\mathbf{24(256\times255)}$ &  \textbf{\xmark} &  \textbf{\xmark} & \textbf{300}\\
    \bottomrule
   \end{tabular}
   \end{center}
\end{table}

\noindent
\textbf{Efficiency Without Global Synchronization: } Unlike prior state-of-the-art SSL methods that rely on embedding concatenation across GPUs (Gather Layer) or global batch normalization across devices, our approach achieves comparable performance without either. This design is more computationally efficient (we refer to Table \ref{table:complexityonImageNet}), requires fewer negative pairs, and enables fully asynchronous training across devices.

\subsection{Optimization Dynamics and Representation Diversity}
\label{ssec:optimization_dynamics}
To better understand the optimization behavior induced by SMI, we analyze the training dynamics of the proposed nonlinear dependency objective throughout self-supervised pretraining. In particular, we study gradient stability, representation diversity, and the alignment--redundancy tradeoff during training. Our analysis reveals that SMI induces substantially different optimization dynamics compared to conventional redundancy reduction objectives such as Barlow Twins, resulting in more stable optimization and improved representation diversity under low-resource training settings.

\begin{figure}[H]
    \centering
    \includegraphics[width=0.7\columnwidth]{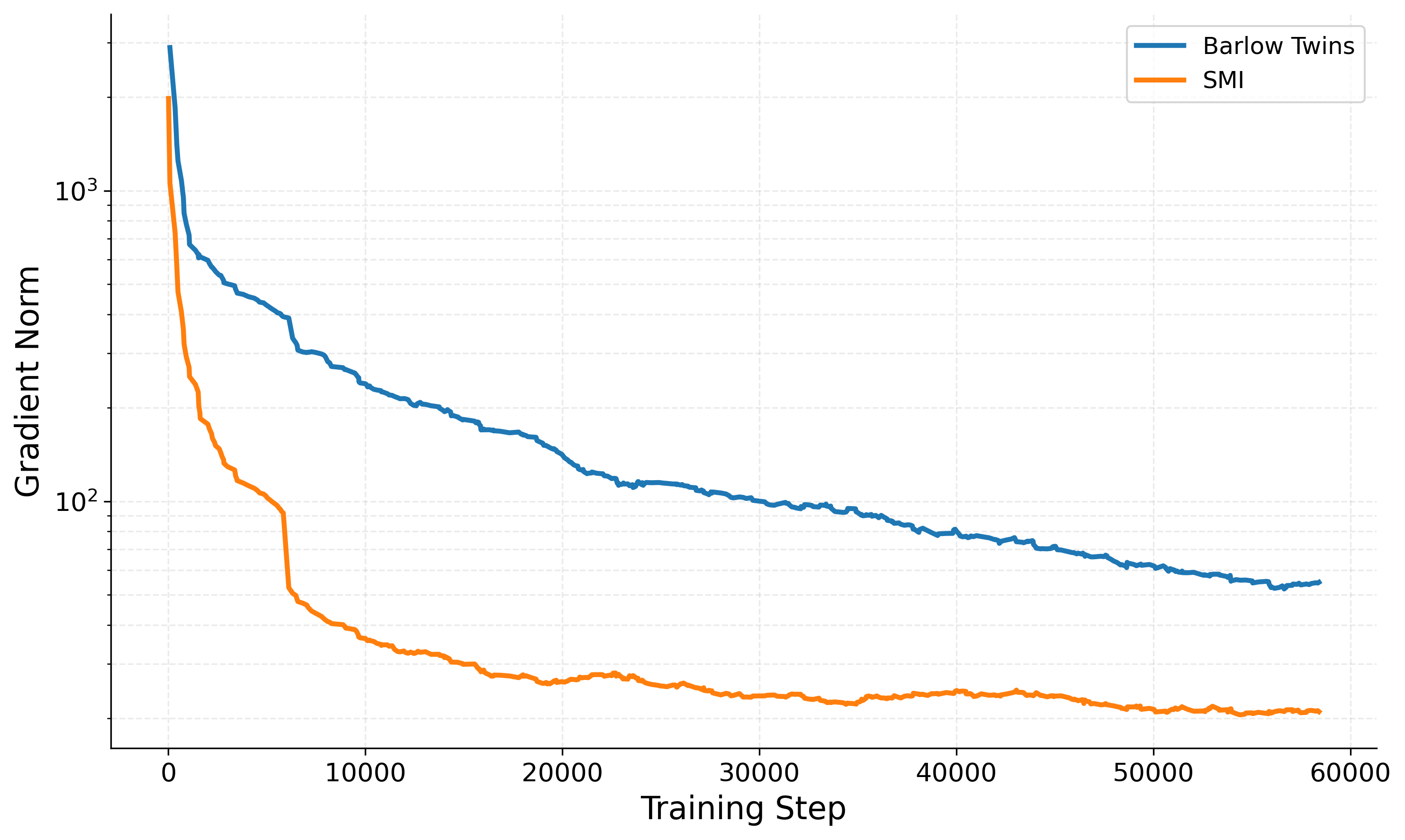}
    \vspace{10pt}
    \caption{Gradient norm evolution during SSL pretraining on CIFAR10 using ResNet50. Curves are smoothed using a rolling average window of 50 steps and visualized on a logarithmic scale. SMI exhibits consistently lower gradient magnitudes and smoother optimization dynamics compared to Barlow Twins.}
    \label{fig:grad_norm_comparison}
\end{figure}

Figure~\ref{fig:grad_norm_comparison} compares the evolution of gradient norms during SSL pretraining on CIFAR10 using a ResNet50 backbone. Across training, SMI consistently exhibits smoother and lower-magnitude gradient dynamics relative to Barlow Twins. In contrast, Barlow Twins maintains significantly larger gradient magnitudes throughout optimization, suggesting a more aggressive redundancy suppression optimization. The smoother optimization trajectory of SMI is consistent with the nonlinear dependency formulation introduced by the logarithmic mutual-information-inspired objective, which adaptively rescales gradient sensitivity based on the correlation structure of the learned representations. These results indicate that SMI produces a more stable optimization landscape while avoiding unstable gradient amplification during training.

\begin{figure}[H]
    \centering
    \includegraphics[width=0.7\columnwidth]{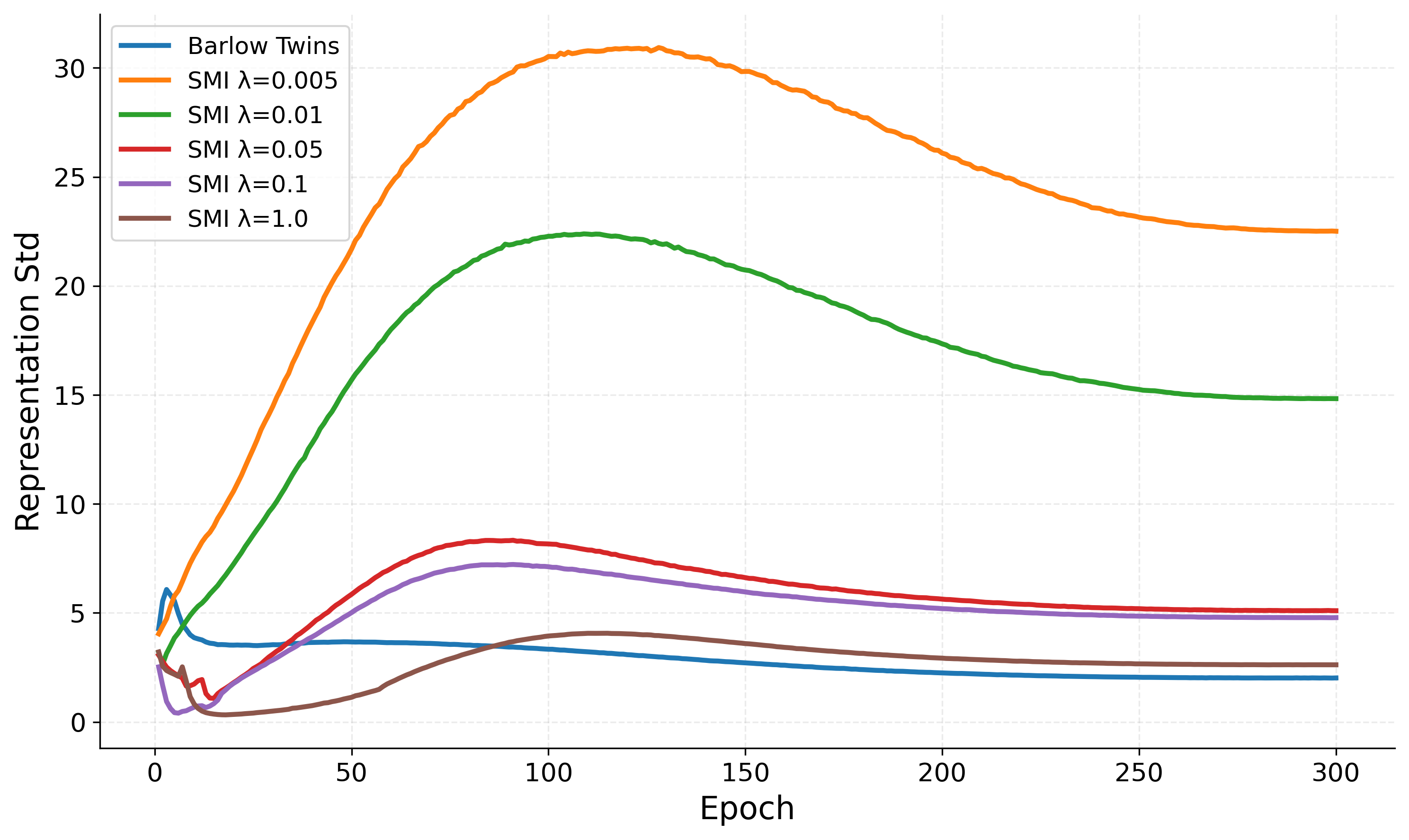}
    \vspace{10pt}
    \caption{Evolution of representation standard deviation during self-supervised training for Barlow Twins and SMI under different regularization strengths $\lambda$ on CIFAR10 using a ResNet50 backbone. SMI enables controllable representation diversity across different $\lambda$ regimes, revealing a non-monotonic relationship between representation diversity and downstream transfer performance. Intermediate diversity regimes ($\lambda = 0.01$) achieve the strongest linear evaluation accuracy, while excessively low or high diversity leads to degraded transfer performance. In contrast, Barlow Twins converges to a comparatively lower-diversity representation regime.
}
    \label{fig:feature_std}
\end{figure}

Beyond optimization stability, we next investigate how SMI influences the diversity of learned representations. Figure~\ref{fig:feature_std} shows the evolution of feature standard deviation during training under different regularization strengths $\lambda$. Unlike Barlow Twins, which converges toward a comparatively low-diversity representation regime, SMI enables controllable representation diversity across different $\lambda$ settings. Interestingly, the relationship between representation diversity and downstream transfer performance is non-monotonic. Moderate diversity regimes achieve the strongest linear evaluation performance, whereas excessively low or excessively high diversity leads to degraded transfer accuracy. In particular, $\lambda = 0.01$ consistently produces the best tradeoff between representation diversity and downstream performance.

\begin{figure}[H]
    \centering
    \includegraphics[width=0.7\columnwidth]{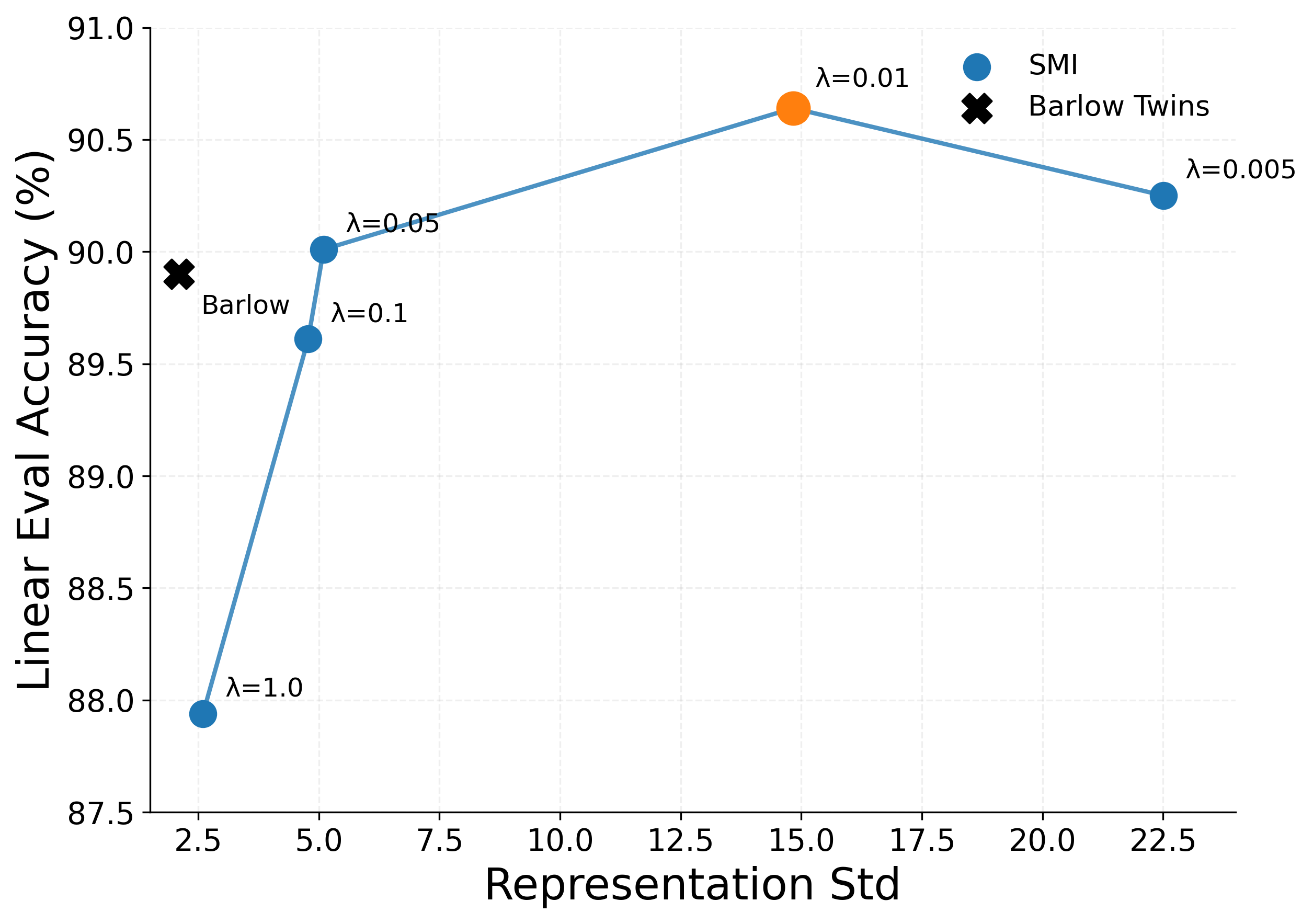}
    \vspace{10pt}
    \caption{Linear evaluation accuracy as a function of representation standard deviation for different SMI regularization strengths $\lambda$ on CIFAR10 using a ResNet50 backbone. SMI enables controllable representation diversity across different $\lambda$ regimes. Performance improves as representation diversity increases up to an intermediate regime ($\lambda = 0.01$), after which excessive diversity leads to a mild degradation in downstream transfer performance. Barlow Twins operates in a comparatively lower-diversity regime.
}
    \label{fig:acc_vs_std}
\end{figure}

\begin{figure}[H]
    \centering
    \includegraphics[width=0.7\columnwidth]{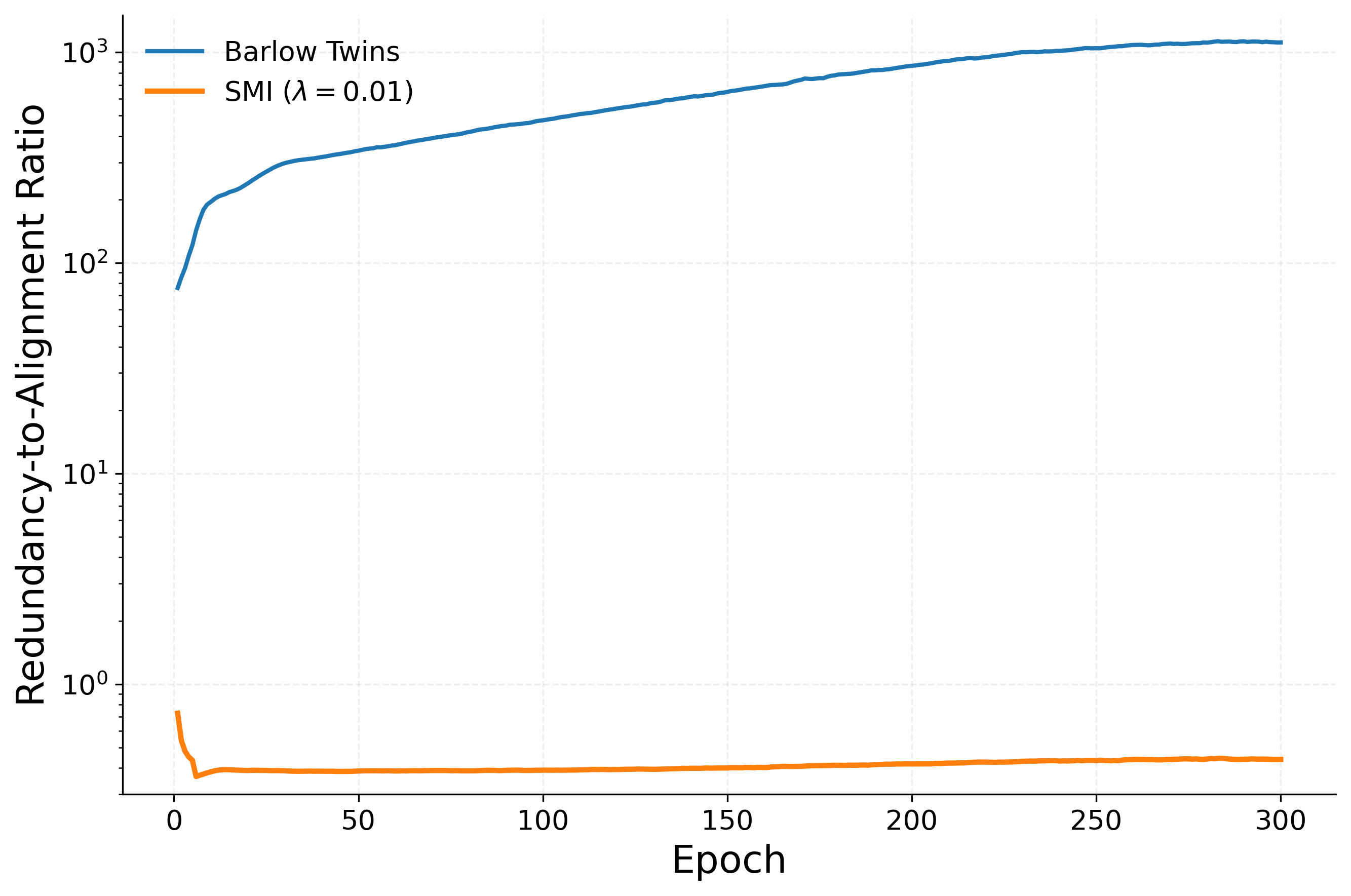}
    \vspace{10pt}
    \caption{Evolution of the redundancy-to-alignment ratio during training for Barlow Twins and SMI ($\lambda = 0.01$) on CIFAR10 using a ResNet50 backbone. Barlow Twins progressively shifts toward a redundancy-dominated optimization regime, whereas SMI maintains a substantially lower and more stable ratio throughout training. This behavior suggests that SMI preserves a more balanced alignment--redundancy tradeoff, consistent with the higher representation diversity and improved downstream transfer performance observed in linear evaluation.
}
    \label{fig:off_diag_vs_on_diag}
\end{figure}

To further analyze this behavior, Figure~\ref{fig:acc_vs_std} plots linear evaluation accuracy as a function of representation standard deviation across different SMI regularization strengths. The results suggest that representation diversity alone is insufficient for optimal transfer performance. Instead, downstream performance appears to depend on maintaining a balanced intermediate diversity regime. Excessively constrained representations reduce feature expressiveness, while overly diverse representations may weaken semantic consistency across views.

Finally, we analyze the alignment--redundancy tradeoff during training. Figure~\ref{fig:off_diag_vs_on_diag} compares the redundancy-to-alignment ratio for SMI and Barlow Twins, defined as the relative contribution of off-diagonal redundancy suppression and positive-pair alignment. While Barlow Twins progressively places greater emphasis on redundancy reduction throughout training, SMI maintains a lower and more stable ratio across all epochs. This suggests that the nonlinear dependency objective of SMI preserves a more balanced tradeoff between alignment and redundancy reduction.

Overall, these results indicate that SMI modifies the optimization geometry of self-supervised learning in a manner fundamentally different from standard correlation matching objectives. Rather than aggressively minimizing redundancy alone, SMI maintains a more balanced alignment–redundancy tradeoff throughout training, enabling controllable representation diversity and improved downstream transfer under computationally constrained settings.

\subsection{Small-Scale Low-Resource Evaluation}
To evaluate the effectiveness of SMI under computationally constrained settings, we compare the proposed objective against Barlow Twins across multiple small-scale benchmarks, including CIFAR10, CIFAR100, ImageNette, and ImageWoof. All methods are pretrained using identical self-supervised training configurations, including the same backbone architecture, augmentation pipeline, optimizer, training schedule, and hyperparameter settings, followed by standard linear evaluation.

\begin{table}[H]
\centering
\caption{
Linear evaluation performance comparison between Barlow Twins and the proposed SMI objective across multiple datasets using a ResNet50 backbone. For fair comparison, all methods are pretrained under identical self-supervised training settings, including the same architecture, optimizer, augmentation pipeline, training schedule, and hyperparameter configuration, followed by evaluation using a frozen linear classifier protocol. SMI consistently improves downstream transfer performance over the Barlow Twins baseline.
}
\vspace{10pt}
\label{tab:linear_eval_results}
\begin{tabular}{lcc}
\toprule

\textbf{Dataset} & \textbf{Barlow Twins} & \textbf{SMI (Ours)} \\

\midrule

CIFAR10 \cite{cifar}     & 89.90 & \textbf{90.64} \\
CIFAR100 \cite{cifar}    & 65.24    & \textbf{66.35} \\
ImageNette \cite{howard2019imagenette}   & 87.77   & \textbf{90.55} \\
Imagewoof \cite{howard2019imagewoof}   & 68.92   & \textbf{75.03} \\

\bottomrule
\end{tabular}

\end{table}

Table~\ref{tab:linear_eval_results} summarizes the downstream linear evaluation performance across all datasets. SMI consistently improves transfer performance relative to Barlow Twins on every benchmark. While moderate gains are observed on CIFAR10 and CIFAR100, larger improvements emerge on the more challenging fine-grained datasets, particularly ImageWoof, where SMI improves Top-1 accuracy by more than 6\%. These results suggest that the proposed nonlinear dependency objective generalizes effectively across diverse low-resource settings and remains particularly robust when fine-grained semantic discrimination is required.

\begin{figure}[H]
    \centering
    \includegraphics[width=0.9\columnwidth]{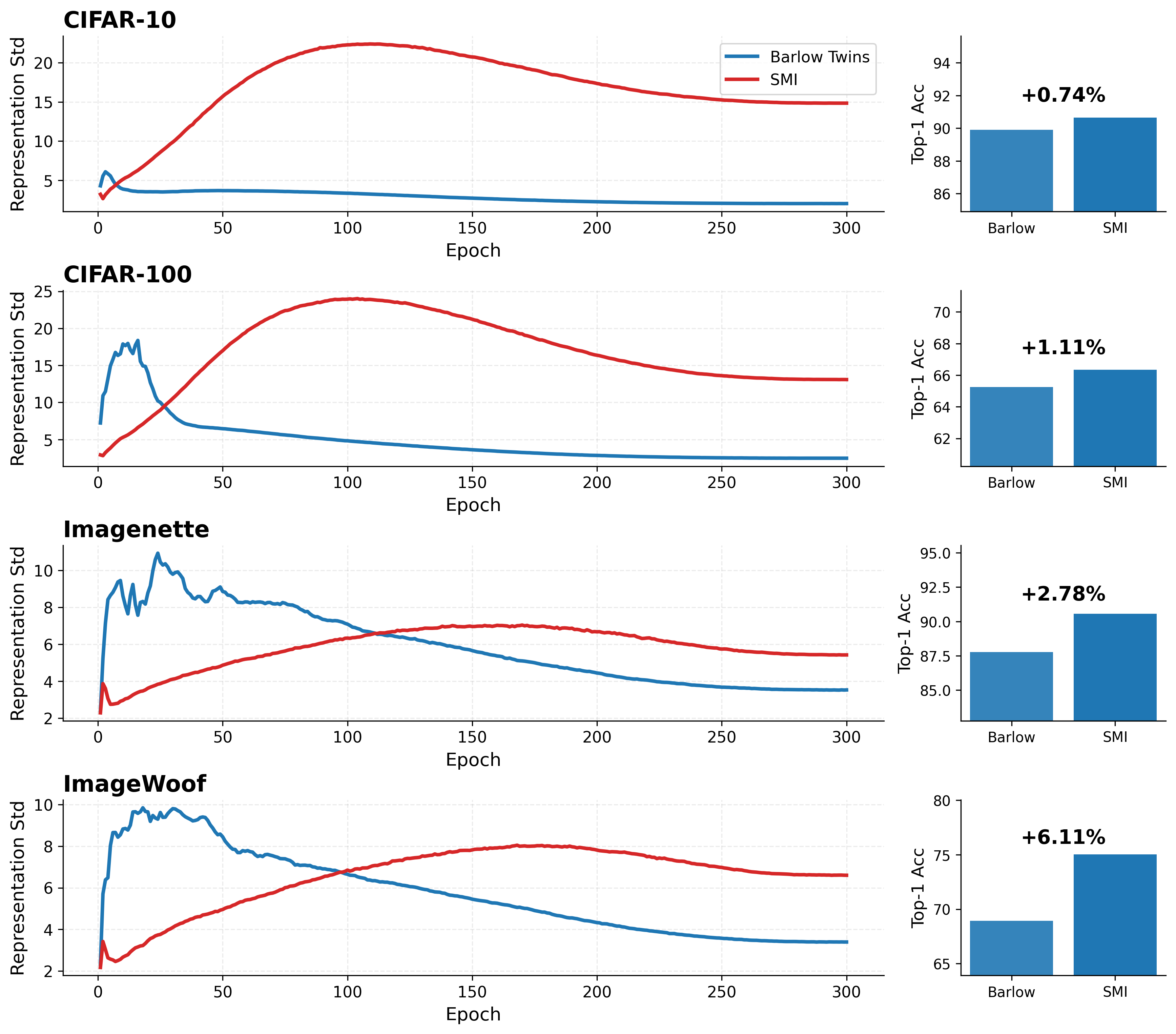}
    \vspace{10pt}
    \caption{Feature standard deviation dynamics during self-supervised pretraining and corresponding downstream linear evaluation performance across CIFAR-10, CIFAR-100, Imagenette, and ImageWoof using a ResNet-50 backbone. For each dataset, the left panel shows the evolution of feature standard deviation throughout training for Barlow Twins and the proposed SMI objective, while the right panel reports the corresponding Top-1 linear evaluation accuracy. Across all datasets, SMI exhibits more stable and consistent representation statistics, correlating with improved downstream classification performance, with the largest gains observed on the fine-grained ImageWoof benchmark.}
    \label{fig:feature_std_summary}
\end{figure}

To further investigate the relationship between representation dynamics and downstream performance, Figure~\ref{fig:feature_std_summary} compares the evolution of feature standard deviation during training across all datasets. Across every benchmark, SMI exhibits substantially more stable and consistent representation statistics compared to Barlow Twins. In particular, SMI avoids the progressive reduction in feature diversity observed in Barlow Twins while maintaining stable representation variance throughout training.

Interestingly, the observed representation dynamics strongly correlate with downstream transfer performance. Datasets where SMI maintains more stable intermediate diversity consistently exhibit improved linear evaluation accuracy. This trend is especially pronounced on ImageWoof, where the largest downstream performance gain coincides with the clearest separation in representation statistics between the two methods. These observations are consistent with the optimization analysis presented in Section~\ref{ssec:smi_vs_corr} and further support the hypothesis that balanced representation diversity plays a critical role in stable self-supervised learning.

Overall, these results demonstrate that SMI provides improved robustness and representation stability under modest computational settings without relying on large batch sizes, momentum encoders, or global synchronization mechanisms. The consistent gains across multiple datasets suggest that the proposed nonlinear dependency objective generalizes effectively beyond large-scale distributed training scenarios.

\subsection{Representation Geometry Analysis}
Beyond downstream performance, we analyze the geometry of the learned representation space through representation diversity, variance concentration, and spatial activation patterns. These analyses complement the optimization dynamics presented in Section~\ref{ssec:optimization_dynamics} and provide additional insight into the observed transfer performance gains.
\subsubsection{Representation Collapse}
\begin{figure}[H]
    \centering
    \subfigure[Eigenspectrum of Representations]{
        \label{fig:eigen_spectrum_cumulative}
        \includegraphics[width=0.42\columnwidth]{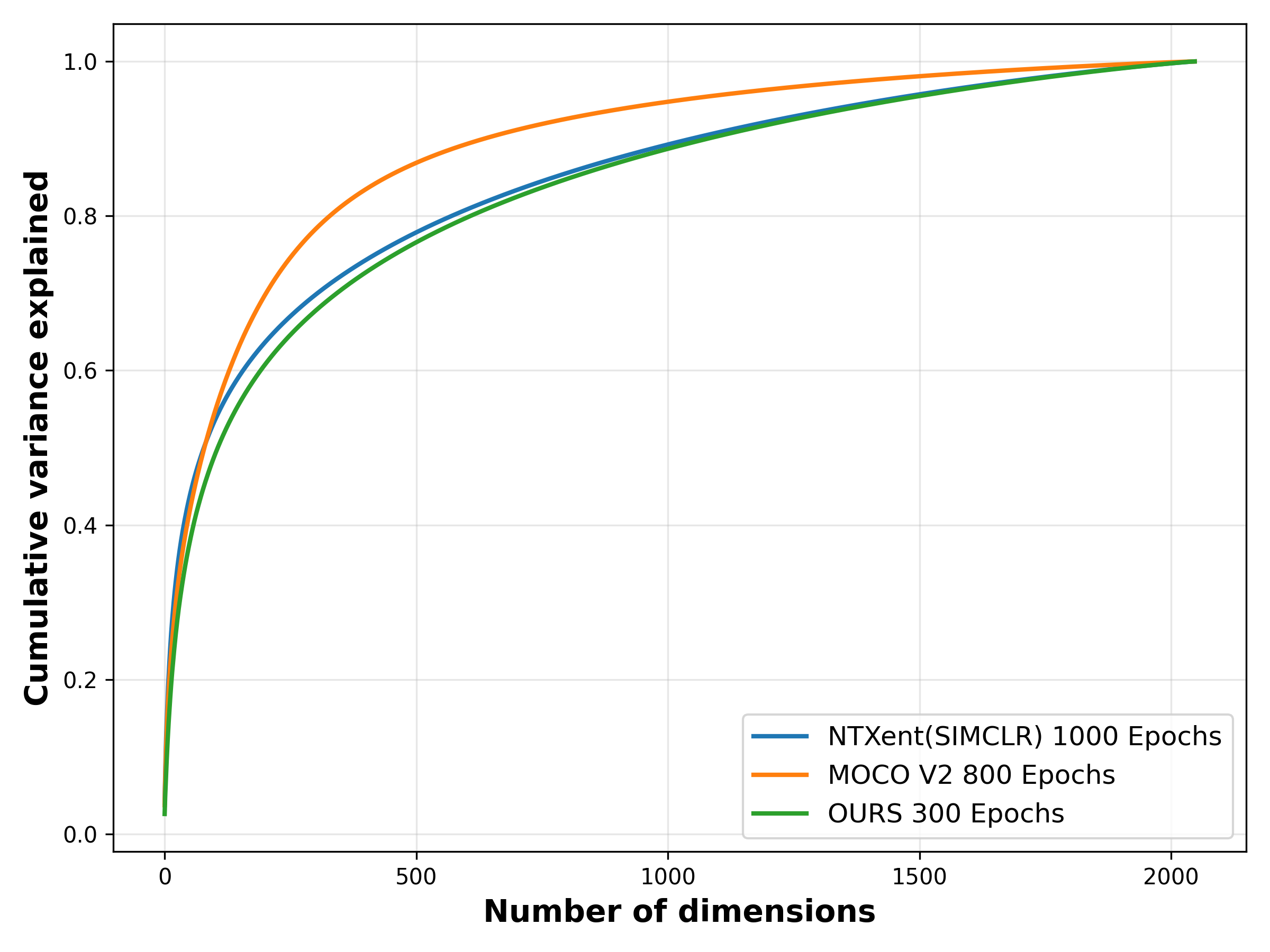}
    }
    \hfill
    \subfigure[Embedding Variance]{
        \label{fig:violinplot}
        \includegraphics[width=0.53\columnwidth]
        {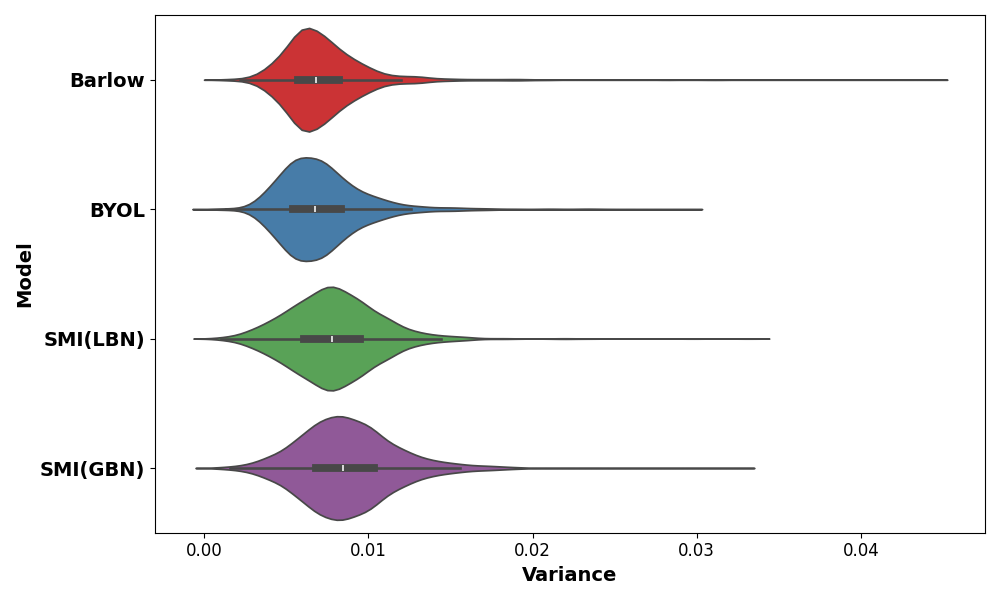}
    }
    \caption{\ref{fig:eigen_spectrum_cumulative} Cumulative explained variance of the representation eigenspectrum. Steeper accumulation indicates stronger variance in a small number of components, commonly associated with representational collapse. \ref{fig:violinplot} Distribution of embedding variance across SSL methods. Higher variance indicates greater representation diversity and reduced collapse.
}
    \label{fig:ablation_rep2}
\end{figure}

To assess representation diversity and potential collapse, we analyze the embedding variance distribution and the cumulative explained variance of the representation eigenspectrum on the ImageNet validation set. Specifically, we compute the cumulative explained variance from the normalized eigenvalues of the feature covariance matrix and compare representations learned by SMI against NT-Xent (SimCLR) and MoCo v2. Figure~\ref{fig:violinplot} and Figure~\ref{fig:eigen_spectrum_cumulative} summarize the resulting variance distributions and eigenspectrum statistics.

\begin{figure}[!htbp]
    \centering
    \includegraphics[width=\textwidth]{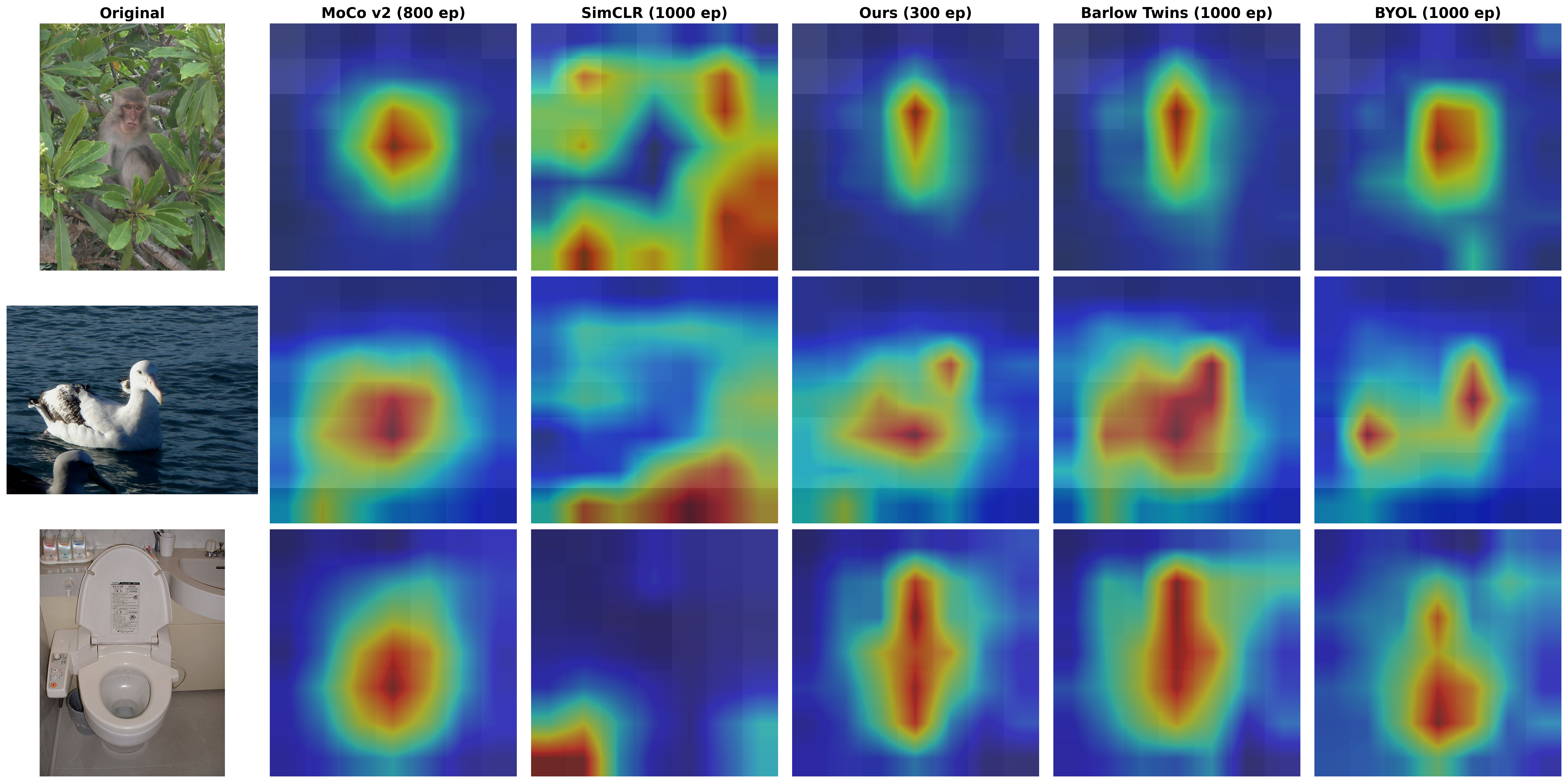}
    \vspace{10pt} 
    \caption{Activation energy maps computed from the spatial $\ell_2$ norm of the final convolutional feature maps of SSL-pretrained ResNet-50 encoders on ImageNet. Compared to NT-Xent (SimCLR), our method produces more spatially concentrated and object-centric responses with reduced background activation. Relative to BYOL and Barlow Twins, SMI exhibits more compact and localized activations, particularly in cluttered scenes.}
    \label{fig:qualitative_analysis}
\end{figure}

\subsubsection{Qualitative Representation Analysis}
To qualitatively assess the spatial focus of the learned representations, we visualize activation energy maps obtained from SSL-pretrained encoders on challenging ImageNet examples. As shown in Figure~\ref{fig:qualitative_analysis}, NT-Xent (SimCLR) produces diffuse activations that often extend into background regions and high-frequency textures, indicating a stronger sensitivity to local appearance cues. MoCo v2 exhibits improved localization relative to SimCLR but retains comparatively broad responses. BYOL and Barlow Twins yield more object-centric activations, though these are often spatially smooth and cover larger object regions. In contrast, our method consistently produces more compact and spatially concentrated activation maps with reduced background activation, particularly in cluttered scenes. These observations are consistent with the improved representation diversity observed in Figures~\ref{fig:eigen_spectrum_cumulative} and \ref{fig:violinplot}.

\section{Conclusions}
We presented Semantic Mutual Information (SMI), a novel self-supervised learning objective based on a Gaussian mutual-information-inspired dependency transformation. Unlike conventional correlation matching approaches, SMI performs alignment and redundancy reduction in a nonlinear dependency space, resulting in correlation-dependent optimization sensitivity and fundamentally different optimization behavior. Our analyses show that these properties lead to smoother optimization dynamics, improved representation diversity, and a more favorable alignment--redundancy balance.

Across multiple benchmarks, SMI consistently improves downstream transfer performance while significantly reducing computational complexity and eliminating the need for Global Batch Normalization or Gather Layers. These results demonstrate that effective self-supervised learning can be achieved through dependency-based objectives without relying on increasingly expensive training protocols. More broadly, this work highlights the potential of MI-inspired dependency optimization as a scalable and computationally efficient direction for self-supervised representation learning.

\clearpage

\bibliography{references}

@inproceedings{poole2019variational,
  title={On variational bounds of mutual information},
  author={Poole, Ben and Ozair, Sherjil and Van Den Oord, Aaron and Alemi, Alex and Tucker, George},
  booktitle={International Conference on Machine Learning},
  pages={5171--5180},
  year={2019},
  organization={PMLR}
}

@inproceedings{belghazi2018mutual,
  title={Mutual information neural estimation},
  author={Belghazi, Mohamed Ishmael and Baratin, Aristide and Rajeshwar, Sai and Ozair, Sherjil and Bengio, Yoshua and Courville, Aaron and Hjelm, Devon},
  booktitle={International conference on machine learning},
  pages={531--540},
  year={2018},
  organization={PMLR}
}

@inproceedings{chen2020simple,
  title={A simple framework for contrastive learning of visual representations},
  author={Chen, Ting and Kornblith, Simon and Norouzi, Mohammad and Hinton, Geoffrey},
  booktitle={International conference on machine learning},
  pages={1597--1607},
  year={2020},
  organization={PMLR}
}

@article{cifar,
  title={Learning multiple layers of features from tiny images},
  author={Krizhevsky, Alex and Hinton, Geoffrey and others},
  journal={Technical Report, University of Toronto},
  year={2009},
  publisher={Toronto, ON, Canada}
}

@inproceedings{stl10,
  title={An analysis of single-layer networks in unsupervised feature learning},
  author={Coates, Adam and Ng, Andrew and Lee, Honglak},
  booktitle={Proceedings of the fourteenth international conference on artificial intelligence and statistics},
  pages={215--223},
  year={2011},
  organization={JMLR Workshop and Conference Proceedings}
}

@article{LARS,
  title={Large batch training of convolutional networks},
  author={You, Yang and Gitman, Igor and Ginsburg, Boris},
  journal={arXiv preprint arXiv:1708.03888},
  year={2017}
}

@article{swav,
  title={Unsupervised learning of visual features by contrasting cluster assignments},
  author={Caron, Mathilde and Misra, Ishan and Mairal, Julien and Goyal, Priya and Bojanowski, Piotr and Joulin, Armand},
  journal={Advances in neural information processing systems},
  volume={33},
  pages={9912--9924},
  year={2020}
}

@inproceedings{imagenet,
  title={Imagenet: A large-scale hierarchical image database},
  author={Deng, Jia and Dong, Wei and Socher, Richard and Li, Li-Jia and Li, Kai and Fei-Fei, Li},
  booktitle={2009 IEEE conference on computer vision and pattern recognition},
  pages={248--255},
  year={2009},
  organization={Ieee}
}

@article{sklar1973random,
  title={Random variables, joint distribution functions, and copulas},
  author={Sklar, Abe},
  journal={Kybernetika},
  volume={9},
  number={6},
  pages={449--460},
  year={1973},
  publisher={Institute of Information Theory and Automation AS CR}
}

@inproceedings{he2020momentum,
  title={Momentum contrast for unsupervised visual representation learning},
  author={He, Kaiming and Fan, Haoqi and Wu, Yuxin and Xie, Saining and Girshick, Ross},
  booktitle={Proceedings of the IEEE/CVF conference on computer vision and pattern recognition},
  pages={9729--9738},
  year={2020}
}

@inproceedings{barlowtwins,
  title={Barlow twins: Self-supervised learning via redundancy reduction},
  author={Zbontar, Jure and Jing, Li and Misra, Ishan and LeCun, Yann and Deny, St{\'e}phane},
  booktitle={International conference on machine learning},
  pages={12310--12320},
  year={2021},
  organization={PMLR}
}

@inproceedings{vicreg,
  title={VICReg: Variance-Invariance-Covariance Regularization For Self-Supervised Learning},
  author={Bardes, Adrien and Ponce, Jean and Lecun, Yann},
  booktitle={ICLR 2022-International Conference on Learning Representations},
  year={2022}
}

@inproceedings{siamsiam,
  title={Exploring simple siamese representation learning},
  author={Chen, Xinlei and He, Kaiming},
  booktitle={Proceedings of the IEEE/CVF conference on computer vision and pattern recognition},
  pages={15750--15758},
  year={2021}
}

@article{byol,
  title={Bootstrap your own latent-a new approach to self-supervised learning},
  author={Grill, Jean-Bastien and Strub, Florian and Altch{\'e}, Florent and Tallec, Corentin and Richemond, Pierre and Buchatskaya, Elena and Doersch, Carl and Avila Pires, Bernardo and Guo, Zhaohan and Gheshlaghi Azar, Mohammad and others},
  journal={Advances in neural information processing systems},
  volume={33},
  pages={21271--21284},
  year={2020}
}

@article{moco_v2,
  title={Improved baselines with momentum contrastive learning},
  author={Chen, Xinlei and Fan, Haoqi and Girshick, Ross and He, Kaiming},
  journal={arXiv preprint arXiv:2003.04297},
  year={2020}
}

@inproceedings{PIRL,
  title={Self-supervised learning of pretext-invariant representations},
  author={Misra, Ishan and Maaten, Laurens van der},
  booktitle={Proceedings of the IEEE/CVF conference on computer vision and pattern recognition},
  pages={6707--6717},
  year={2020}
}

@inproceedings{dino,
  title={Emerging properties in self-supervised vision transformers},
  author={Caron, Mathilde and Touvron, Hugo and Misra, Ishan and J{\'e}gou, Herv{\'e} and Mairal, Julien and Bojanowski, Piotr and Joulin, Armand},
  booktitle={Proceedings of the IEEE/CVF international conference on computer vision},
  pages={9650--9660},
  year={2021}
}

@misc{howard2019imagenette,
  title={Imagenette: A smaller subset of 10 easily classified classes from ImageNet},
  author = {Fastai},
  year={2019},
  url={https://github.com/fastai/imagenette}
}

@misc{howard2019imagewoof,
  title={Imagewoof: A subset of 10 classes from ImageNet that are hard to classify},
  author = {Fastai},
  year={2019},
  url={https://github.com/fastai/imagenette}
}

@inproceedings{trim,
  title={Trim: A self-supervised video summarization framework maximizing temporal relative information and representativeness},
  author={Mishra, Pritam and Ballester, Coloma and Karatzas, Dimosthenis},
  booktitle={ICASSP 2026-2026 IEEE International Conference on Acoustics, Speech and Signal Processing (ICASSP)},
  pages={10717--10721},
  year={2026},
  organization={IEEE}
}

@article{trimmer,
  title={TRIMMER: A New Paradigm for Video Summarization through Self-Supervised Reinforcement Learning},
  author={Mishra, Pritam and Ballester, Coloma and Karatzas, Dimosthenis},
  journal={arXiv preprint arXiv:2605.01659},
  year={2026}
}

\clearpage

\appendix
\section{Appendix}
\subsection{Implementation Details}

\subsubsection{Large-scale Training}
\noindent\textbf{Image Augmentations: }
Our image augmentation pipeline consists of the following transformations with corresponding probability values: random cropping, resizing to 224 × 224, horizontal flipping ($p = 0.5$, where $p$ is the probability of applying the transformation during data augmentation), color jittering ([brightness=0.4, contrast=0.4, saturation=0.2, hue=0.1], \(p=0.8\)), converting to grayscale (\(p=0.2\)), Gaussian blurring (\(p=1.0, \forall x^1_i\); \(p=0.1, \forall x^2_i\)), and solarization (\(p=0, \forall x^1_i\); \(p=1.0, \forall x^2_i\)) similar to \cite{byol,barlowtwins}.\\
\textbf{Architecture: }
Building on previous SOTA methods, we use ResNet-50 as our encoder, removing the final classification layer. For the projection head, we adopt the same architecture as Barlow Twins \cite{barlowtwins}: three linear layers with 8192 units each, along with batch normalization and ReLU activation.\\ 
\textbf{Optimization: }
We followed the optimization protocol described in previous SOTA works \cite{byol,barlowtwins,vicreg}, with minor adjustments. Our model was trained with an effective batch size of 2048 for 300 epochs with LARS optimizer \cite{LARS} without Gather Layer or Global Batch normalization across distributed devices. The base learning rate was set to scale with the effective batch size as $\text{batchsize}/256$ for 300 epochs. We adopt the same learning rate for bias, batch normalization, and weight decay parameters as used in Barlow Twins \cite{barlowtwins}. 

\subsubsection{Training on Small Datasets}

\noindent\textbf{Data Augmentation: }
For self-supervised pretraining, two augmented views are generated from each image using a stochastic augmentation pipeline. CIFAR-10 and CIFAR-100 employ RandomResizedCrop (scale $(0.2,1.0)$), RandomHorizontalFlip, ColorJitter (brightness=0.4, contrast=0.4, saturation=0.2, hue=0.1, $p=0.8$), RandomGrayscale ($p=0.2$), and GaussianBlur. For ImageNette and ImageWoof, we adopt a stronger augmentation pipeline consisting of RandomResizedCrop (scale $(0.08,1.0)$), RandomHorizontalFlip, ColorJitter (brightness=0.4, contrast=0.4, saturation=0.4, hue=0.1, $p=0.8$), RandomGrayscale ($p=0.2$), GaussianBlur, and RandomSolarize ($p=0.2$). Detailed parameters for augmentations are provided in Table~\ref{tab:ssl_aug}.

\vspace{3pt}
\noindent\textbf{Architecture: }
All experiments use a ResNet-50 backbone. For CIFAR-10 and CIFAR-100, the standard ImageNet stem is replaced with a ($3\times3$) convolution with stride 1 and padding 1, and the initial max-pooling layer is removed to better accommodate low-resolution inputs. For ImageNette and ImageWoof, the standard ResNet-50 architecture is used. Following Barlow Twins, the encoder output is passed through a three-layer projection head with 4096 hidden units per layer, batch normalization, and ReLU activations.

\vspace{3pt}
\noindent\textbf{Optimization: }
To ensure a consistent low-resource evaluation protocol, identical optimization settings are used across CIFAR-10, CIFAR-100, ImageNette, and ImageWoof. All models are trained using the AdamW optimizer with the same learning-rate ($1e^{-3}$) schedule (cosine annealing), weight decay ($1e^{-4}$), batch size (256), and training duration (300 epochs).

\begin{table}[h]
\centering
\caption{Dataset-specific SSL augmentation settings.}
\label{tab:ssl_aug}
\begin{tabular}{lccc}
\toprule
Dataset & Resolution & Crop Scale & Solarization \\
\midrule
CIFAR-10      & 32  & (0.2,1.0) & No \\
CIFAR-100     & 32  & (0.2,1.0) & No \\
ImageNette    & 160 & (0.08,1.0) & 0.2 \\
ImageWoof     & 160 & (0.08,1.0) & 0.2 \\
\bottomrule
\end{tabular}
\end{table}

\subsection{Derivation of the Gaussian Mutual Information Dependency Measure}
\subsubsection{Sklar's Theorem}

Sklar's theorem states that any joint cumulative distribution function (CDF) can be decomposed into its marginal distributions and a copula that captures the dependency structure between variables. Given random variables $X$ and $Y$ with joint CDF $F_{X,Y}$,

\begin{equation}
F_{X,Y}(x,y) = C(F_X(x),F_Y(y))
\end{equation}

where $F_X$ and $F_Y$ are the marginal CDFs and $C$ denotes the copula function. The corresponding joint density can be written as

\begin{equation}
f_{X,Y}(x,y) = c(F_X(x),F_Y(y))f_X(x)f_Y(y)
\end{equation}

where $c$ denotes the copula density.

\subsubsection{Gaussian Copula}

The Gaussian copula models dependencies using a multivariate Gaussian structure while allowing arbitrary marginal distributions. It is defined as

\begin{equation}
C_{\mathrm{Gaussian}}(u_1,u_2;\rho) =\Phi_\rho\left(\Phi^{-1}(u_1),\Phi^{-1}(u_2)\right)
\end{equation}

where $\Phi_\rho$ denotes the bivariate Gaussian CDF with correlation coefficient $\rho$ and $\Phi^{-1}$ is the inverse standard Gaussian CDF.

Under the Gaussian copula assumption, dependency is fully characterized by the Pearson correlation coefficient $\rho$.

\subsubsection{Mutual Information and Pearson Correlation}

For two continuous random variables, mutual information is defined as

\begin{equation}
I(X;Y) = h(X)+h(Y)-h(X,Y),
\end{equation}

where $h(\cdot)$ denotes differential entropy.

For jointly Gaussian variables with variance $\sigma^2$,

\begin{equation}
h(X) = h(Y) = \frac12 \log(2\pi e \sigma^2)
\end{equation}

The joint entropy is

\begin{equation}
h(X,Y) =\frac12\log\left((2\pi e)^2|\Sigma|\right)
\end{equation}

where $\Sigma$ denotes the covariance matrix. For a bivariate Gaussian distribution,

\begin{equation}
|\Sigma| = \sigma^4(1-\rho^2)
\end{equation}

Substituting into the entropy expression yields

\begin{equation}
h(X,Y) = \log(2\pi e \sigma^2) + \frac12 \log(1-\rho^2).
\end{equation}

Finally,

\begin{equation}
I(X;Y) = -\frac12 \log(1-\rho^2).
\end{equation}

To avoid numerical instability when $\rho \rightarrow 1$, we introduce a small positive constant $\epsilon$ and obtain the dependency measure used throughout this work:

\begin{equation}
I(X;Y) = -\frac12 \log(1-\rho^2+\epsilon).
\end{equation}

\subsection{Additional Optimization Dynamics Analysis}
\begin{figure*}[t]
\centering


\subfigure[CIFAR-100 Gradient Norm]{
\includegraphics[width=0.47\textwidth]{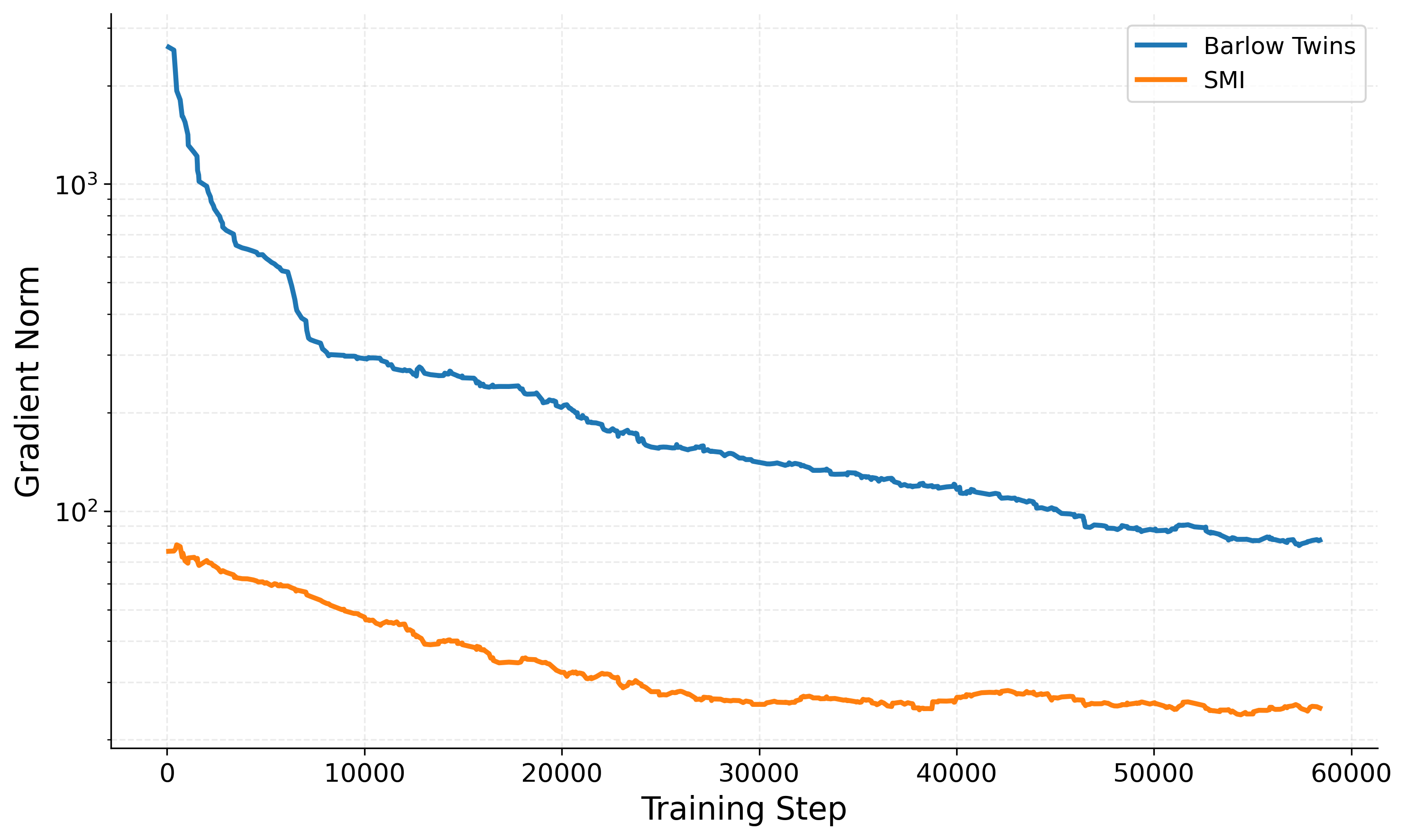}
}
\hfill
\subfigure[CIFAR-100 Train/Val Acc]{
\includegraphics[width=0.47\textwidth]{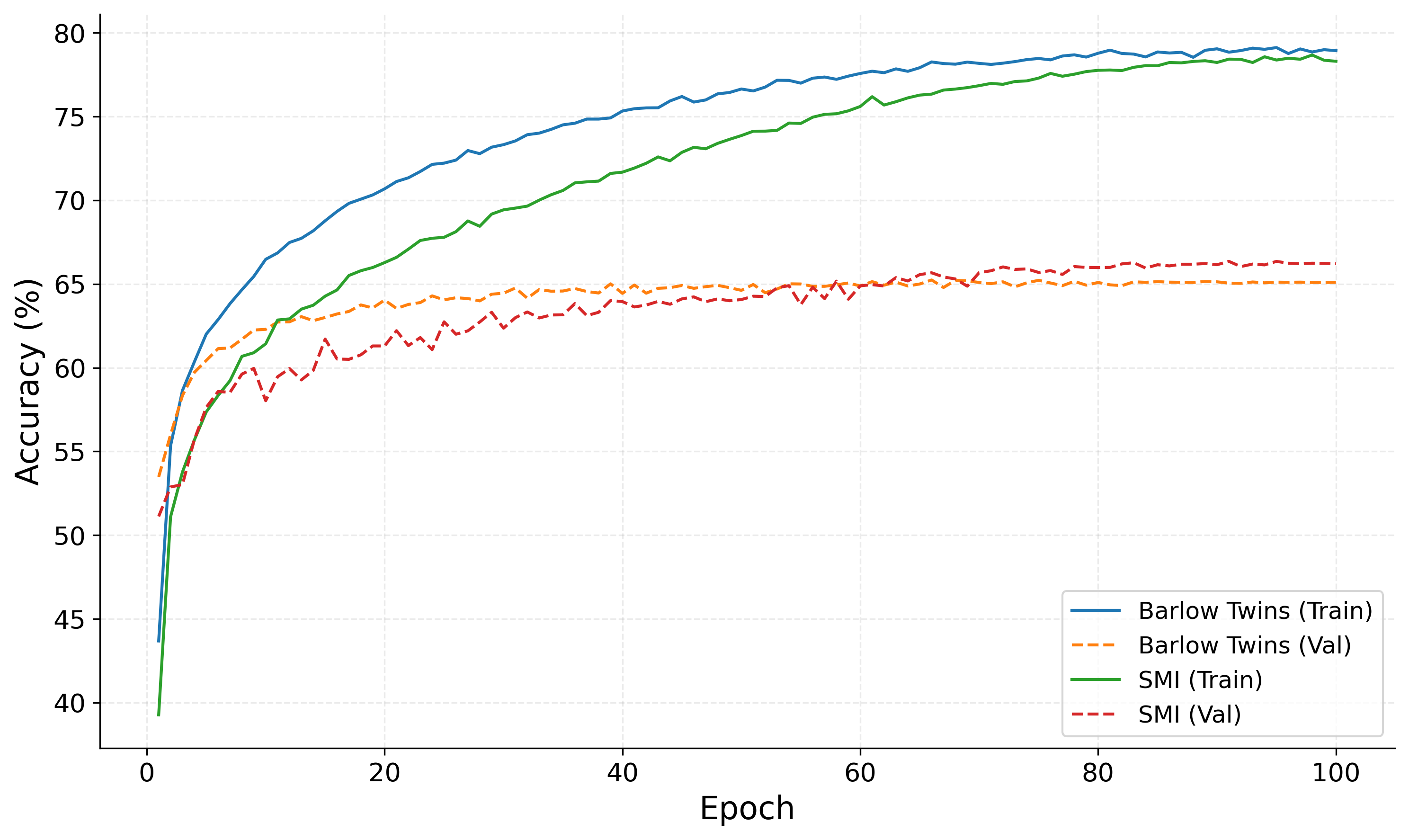}
}

\vspace{8pt}


\subfigure[ImageNette Gradient Norm]{
\includegraphics[width=0.47\textwidth]{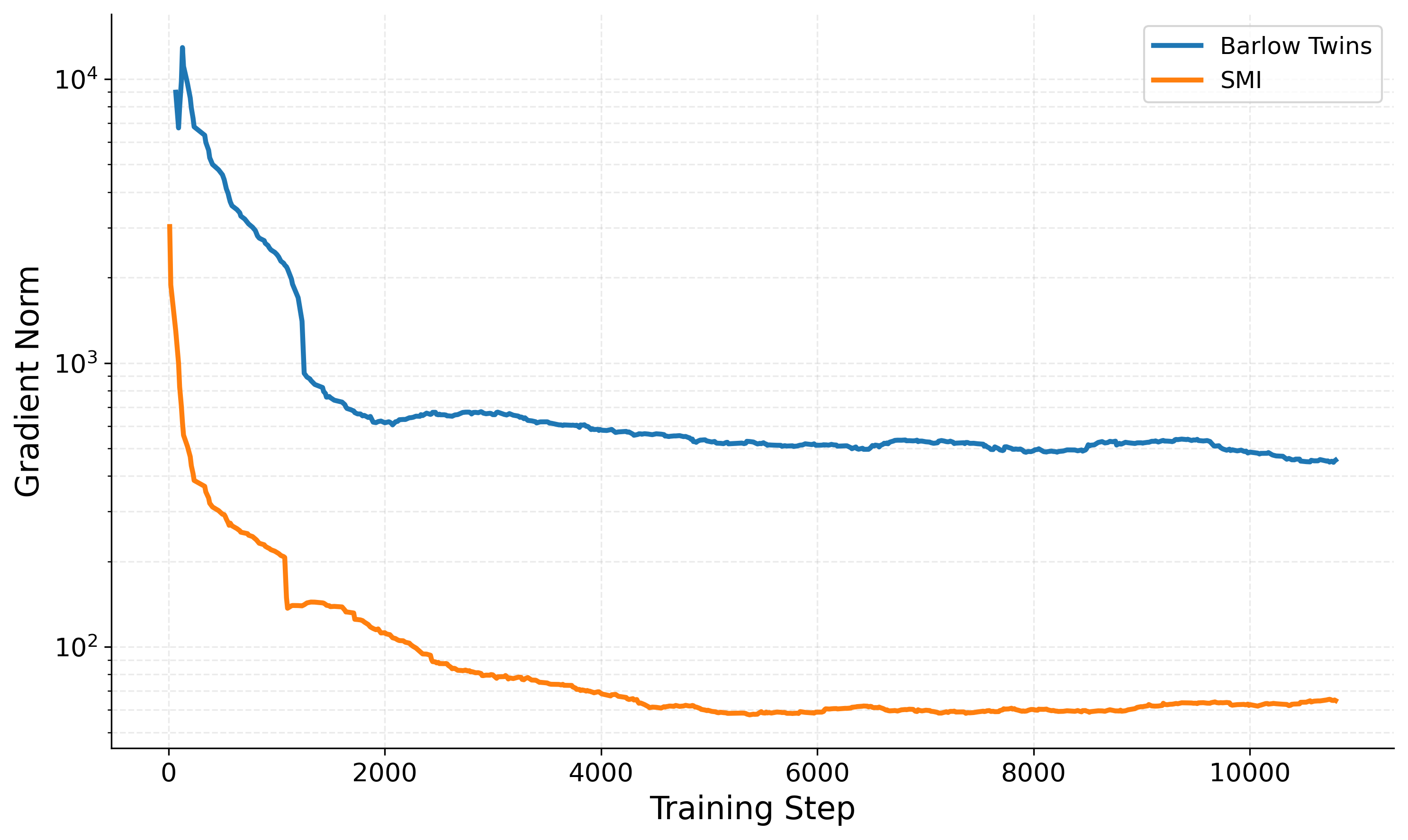}
}
\hfill
\subfigure[ImageNette Train/Val Acc]{
\includegraphics[width=0.47\textwidth]{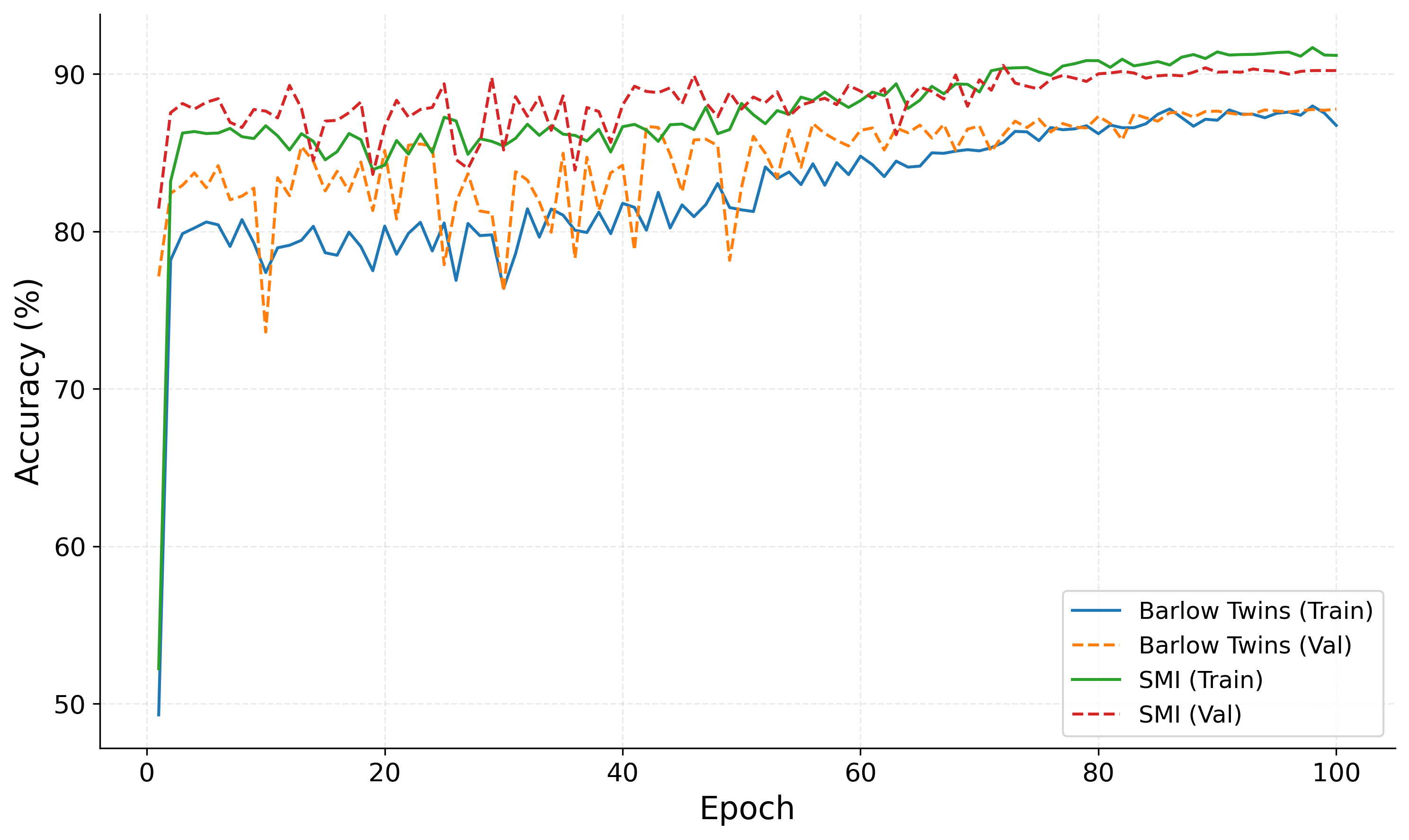}
}

\vspace{8pt}


\subfigure[ImageWoof Gradient Norm]{
\includegraphics[width=0.47\textwidth]{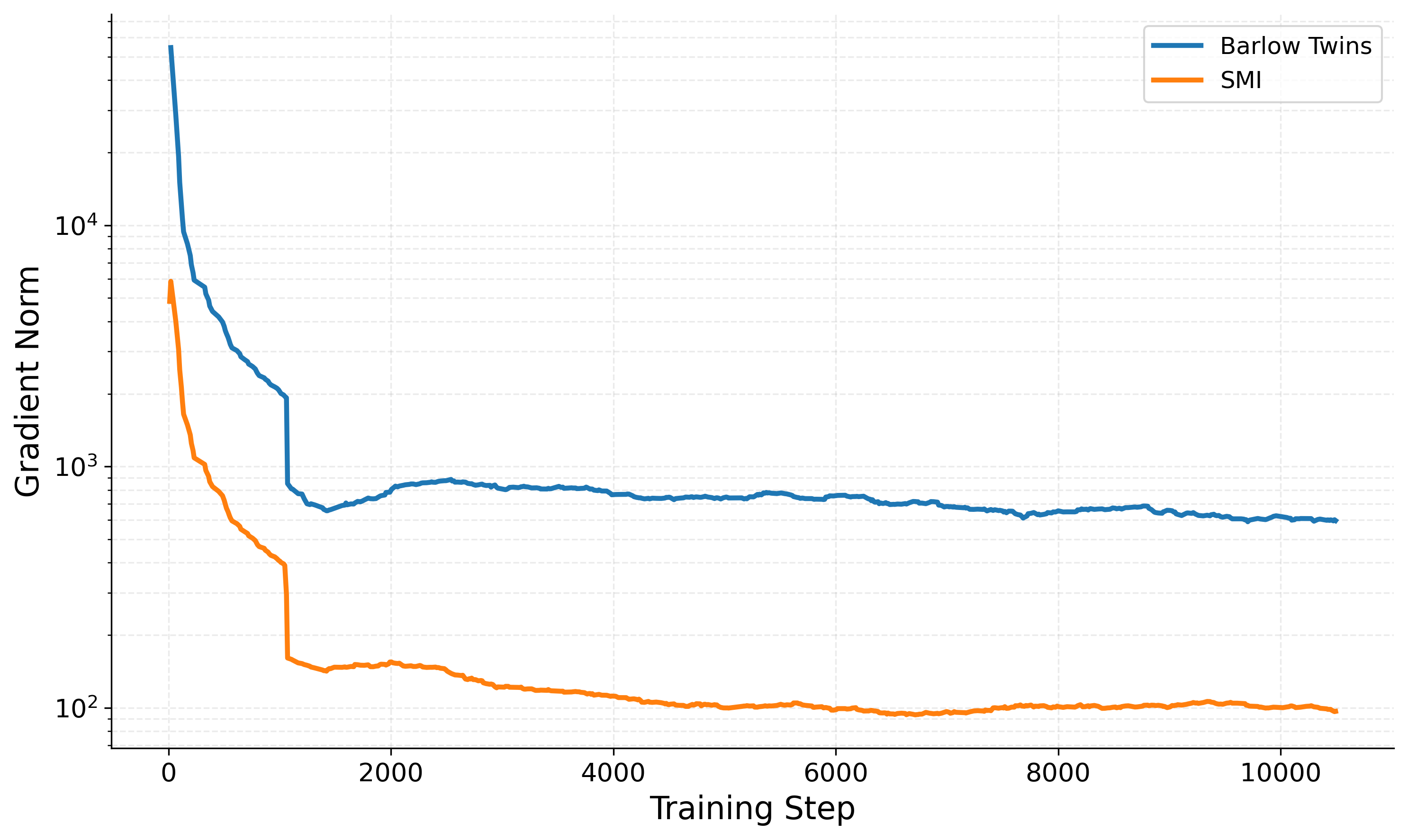}
}
\hfill
\subfigure[ImageWoof Train/Val Acc]{
\includegraphics[width=0.47\textwidth]{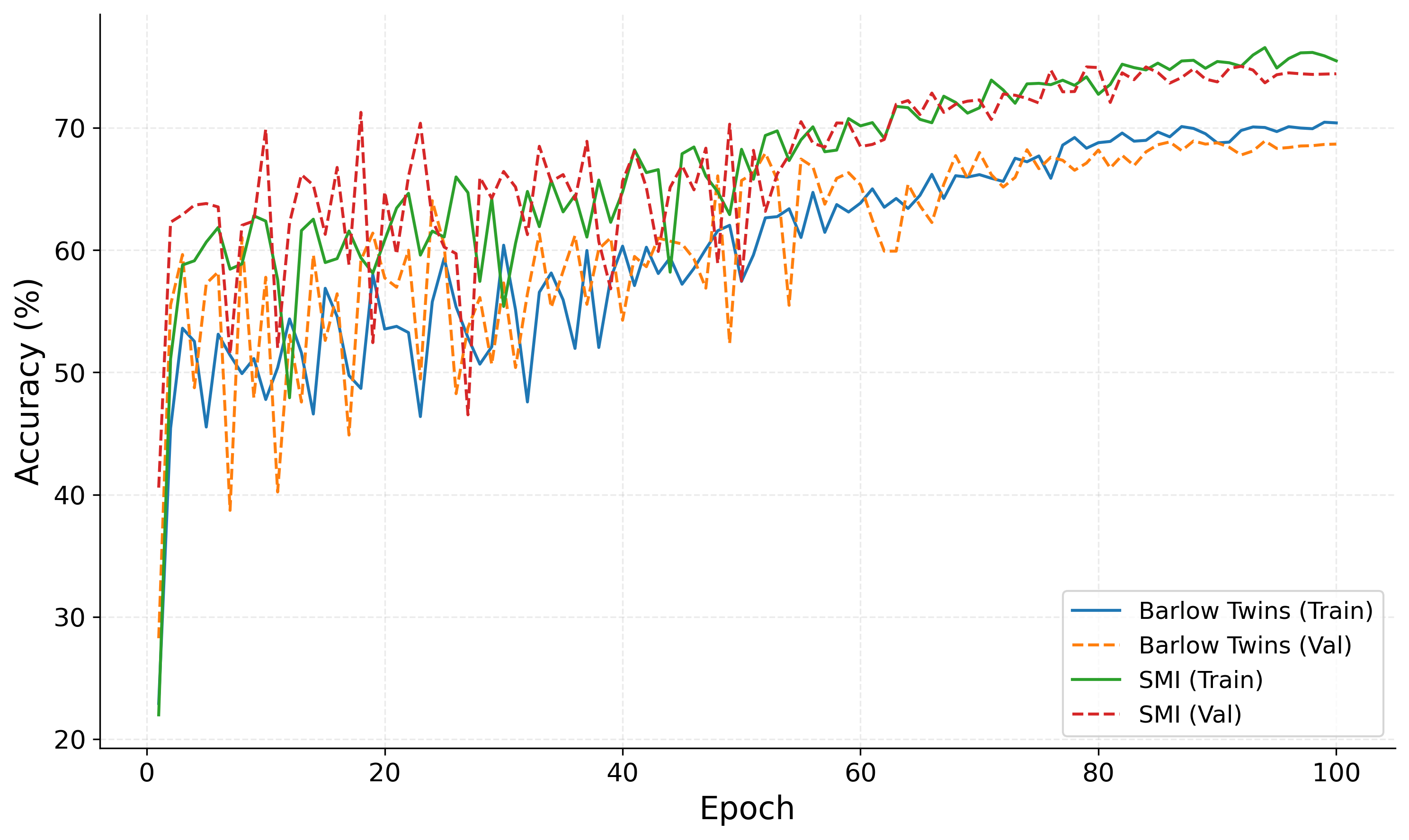}
}

\caption{
Additional training dynamics for the low-resource benchmarks. For each dataset, the left column shows the evolution of gradient norms during SSL pretraining, while the right column reports train and validation accuracy during linear evaluation. Across datasets, SMI exhibits smoother optimization dynamics and consistently improved validation performance relative to the Barlow Twins baseline.
}
\label{fig:supp_training_dynamics}
\end{figure*}

Figure~\ref{fig:supp_training_dynamics} provides additional optimization statistics for CIFAR-100, ImageNette, and ImageWoof. For each dataset, we compare the evolution of gradient norms during self-supervised pretraining and the corresponding train/validation accuracy curves obtained during linear evaluation.

Across all datasets, SMI consistently exhibits lower gradient magnitudes throughout training compared to Barlow Twins, indicating a smoother optimization trajectory. This behavior is particularly pronounced on ImageNette and ImageWoof, where SMI rapidly converges to a stable low-gradient solution while maintaining improved downstream performance.

The train and validation accuracy curves further demonstrate that the improved linear evaluation performance of SMI is not driven by overfitting. In all cases, validation accuracy closely tracks training accuracy throughout optimization, while consistently exceeding the performance achieved by the Barlow Twins baseline. These observations are consistent with the optimization dynamics analysis presented in Section~3.3 of the main paper and further support the hypothesis that the nonlinear dependency objective induces more stable optimization behavior.

\subsection{Additional Representation Geometry}

\begin{figure}[!htbp]
    \centering
    \subfigure[Barlow Twins]{%
        \label{fig:barlow}%
        \includegraphics[width=0.4\columnwidth]{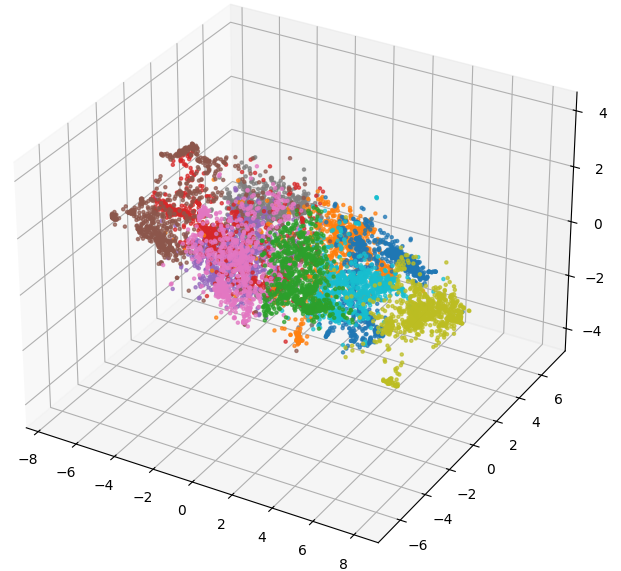}%
    }%
    \subfigure[BYOL]{%
        \label{fig:byol}%
        \includegraphics[width=0.4\columnwidth]{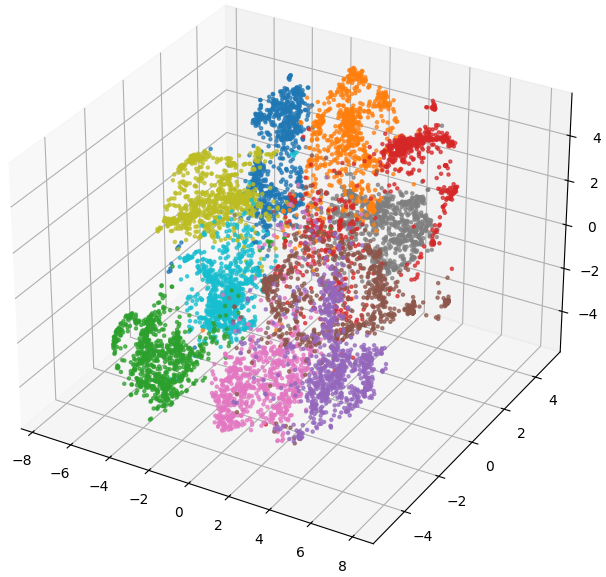}%
    }%
    \includegraphics[width=0.10\columnwidth]{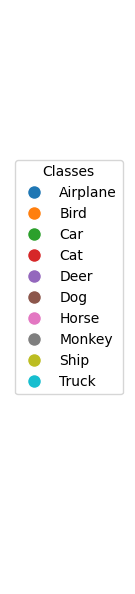}%
    
    \vspace{0.2em} 

    \subfigure[VICREG]{%
        \label{fig:vicreg}%
        \includegraphics[width=0.4\columnwidth]{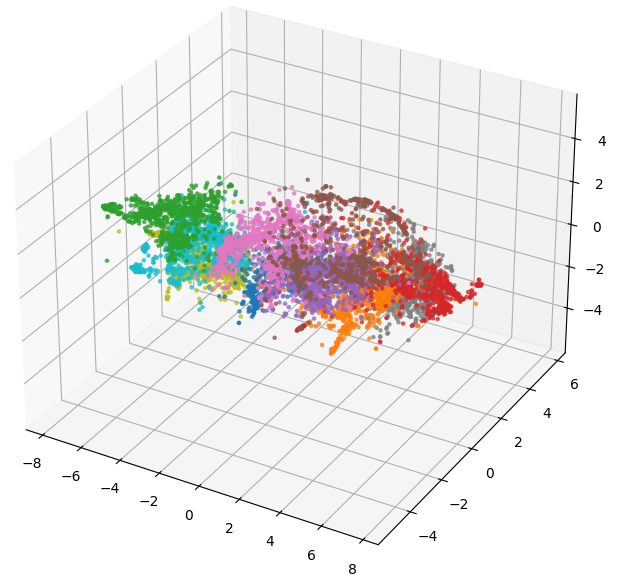}%
    }%
    \subfigure[SMI(Ours)]{%
        \label{fig:ours}%
        \includegraphics[width=0.4\columnwidth]{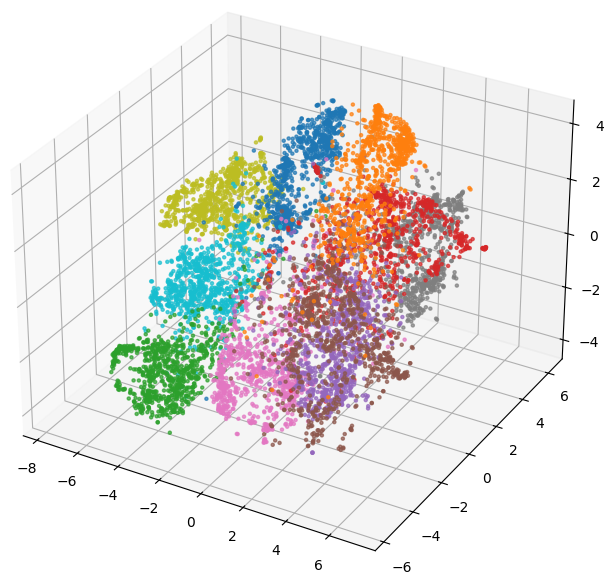}%
    }%
    \makebox[0.08\columnwidth]{}%

    \caption{3D t-SNE visualization of image embeddings from frozen ResNet50 backbone pre-trained on ImageNet and evaluated on the STL10 \cite{stl10} test set (details in the text).}
    \label{fig:3d_tsne_ours}
\end{figure}

To further visualize the structure of the learned representation space, we evaluate the embedding quality of SSL-pretrained models on the STL10~\cite{stl10} test set using 3D t-SNE projections. Figure~\ref{fig:3d_tsne_ours} presents embeddings extracted from frozen ResNet-50 encoders pretrained with different SSL objectives. While t-SNE does not provide a quantitative evaluation of representation quality, it offers an intuitive visualization of cluster compactness and class separability in the learned feature space.

\end{document}